\documentclass{article}

\usepackage{amsmath,amsfonts,bm}

\def\eqref#1{equation~\ref{#1}}

\def\1{\bm{1}}

\DeclareMathAlphabet{\mathsfit}{\encodingdefault}{\sfdefault}{m}{sl}
\SetMathAlphabet{\mathsfit}{bold}{\encodingdefault}{\sfdefault}{bx}{n}

\usepackage{microtype}
\usepackage{graphicx}
\usepackage{subfigure}
\usepackage{booktabs} 
\usepackage{wrapfig}
\usepackage{comment}
\usepackage{hyperref}

\newtheorem{theorem}{Theorem}[section]
\newtheorem{lemma}{Lemma}[section]
\newtheorem{corollary}{Corollary}[section]

\usepackage{hyperref}

\usepackage[accepted]{icml2020}

\icmltitlerunning{AdaX: Adaptive Gradient Descent with Exponential Long Term Memory}

\begin{document}

\twocolumn[
\icmltitle{AdaX: Adaptive Gradient Descent with Exponential Long Term Memory}



\icmlsetsymbol{equal}{*}

\begin{icmlauthorlist}
\icmlauthor{Wenjie Li}{pu,sensetime}
\icmlauthor{Zhaoyang Zhang}{sensetime,mmlab}
\icmlauthor{Xinjiang Wang}{sensetime}
\icmlauthor{Ping Luo}{sensetime,hku}
\end{icmlauthorlist}

\icmlaffiliation{pu}{Department of Statistics, Purdue University.}
\icmlaffiliation{sensetime}{Sensetime Research.}
\icmlaffiliation{mmlab}{Multimedia Lab, Chinese University of Hong Kong.}
\icmlaffiliation{hku}{Department of Computer Science, University of Hong Kong}

\icmlcorrespondingauthor{Ping Luo}{pluo.lhi@gmail.com}

\icmlkeywords{Adam, AdaX, Optimization Algorithm}

\vskip 0.3in
]



\printAffiliationsAndNotice{}  

\begin{abstract}

Although adaptive optimization algorithms such as Adam show fast convergence in many machine learning tasks, this paper identifies a problem of Adam by analyzing its performance in a simple non-convex synthetic problem, showing that Adam's fast convergence would possibly lead the algorithm to local minimums. To address this problem, we improve Adam by proposing a novel adaptive gradient descent algorithm named AdaX. Unlike Adam that ignores the past gradients, AdaX exponentially accumulates the long-term gradient information in the past during training, to adaptively tune the learning rate. We thoroughly prove the convergence of AdaX in both the convex and non-convex settings. Extensive experiments show that AdaX outperforms Adam in various tasks of computer vision and natural language processing and can catch up with Stochastic Gradient Descent.
\end{abstract}

\section{Introduction}
\label{sec: introduction}

Stochastic Gradient Descent (SGD), though proposed in the last century, remains one of the most effective algorithms in training deep neural networks \citep{Robbins1951A}. Many methods have been proposed to accelerate the training process and boost the performance of SGD, such as momentum \citep{Polyak1964Some} and Nesterov's acceleration \citep{Nesterov1983A}. 
Recently, adaptive optimization methods have become popular as they adjust parameters' learning rates in different scales instead of directly controlling the overall step sizes, resulting in smoother training process and faster convergence. For example, AdaGrad \citep{Duchi2011Adaptive} schedules the adaptive learning rate by dividing the gradients by a denominator, which is the square root of the global average of the past gradient squares. It is shown that when the gradients are sparse, AdaGrad can converge faster than vanilla SGD \citep{Duchi2011Adaptive}. However, its generalization performance is limited. \citep{Reddi2018On}. 

In particular, AdaGrad's failure originates from its global average design in the denominator, which increases rapidly when large or dense gradients exist and makes the update steps very small. To address such an issue, other adaptive algorithms have been proposed to replace the denominator by the square root of the exponential moving average of the past gradient squares, such as RMSProp \citep{Tieleman2012RMSProp}, AdaDelta \citep{Zeiler2012AdaDelta}, and Adam \citep{Kingma2015Adam}. 

Among all the above variants, Adam, due to its fast convergence rate and good performances, becomes popular in applications. 
However, 
recent theories have  shown that Adam suffers from non-convergence issues and weak generalization ability \citep{Wilson2017The,Reddi2018On}.
For instance, \citet{Reddi2018On} thoroughly proved that Adam did not guarantee convergence even in a simple convex optimization problem. 
\citet{Shazeer2018AdaFactor} also empirically showed that Adam's parameter updates were not stable and its second moment could be out of date. \citet{Luo2019Adaptive} examined the effective update steps of Adam in training and found that its second moment would produce extreme learning rates. 
\citet{zaheer2019adaptive} found that Adam's performance could be affected by different values of $\epsilon$, which was originally designed to avoid zeros in the denominator.
All the above analysis show that Adam's exponential moving average design is problematic.

This paper addresses the above issues by proposing a novel adaptive gradient descent algorithm, named AdaX, which improves Adam both theoretically and empirically.
The main \textbf{contributions} of this work are three-fold. 

(1) We examine the design of Adam more carefully by changing the convex counterexample in \citet{Reddi2018On} to a more practical setting. We theoretically prove how the second moment in Adam always leads the optimization process to a sub-optimal point even without noisy gradients, revealing that Adam's fast convergence can impair its performances. We also show how AMSGrad \citep{Reddi2018On}, a popular extension of Adam, is unable to solve Adam's problem completely because it doesn't change the exponential moving average design in Adam and its effectiveness relies heavily on the magnitude of the maximal second moment. 

(2) Our new adaptive method AdaX can completely eliminate Adam's problem by replacing the exponential moving average with an exponential long-term memory design as the second moment.
We theoretically prove AdaX gets rid of the second moment instability and the non-convergence issues, and it converges with a speed similar to AMSGrad. 

(3) Extensive experiments show that AdaX outperforms Adam in many tasks of computer vision and natural language processing, such as image recognition on CIFAR-10 \citep{Cifar} and ImageNet \citep{Deng2009Imagenet}, semantic image segmentation on PASCAL VOC2012 \citep{VOC2012}, and language modeling on One Billion Word \citep{chelba2013one} dataset. Moreover, AdaX's performance can catch up with SGD with momentum with a much faster convergence, which no other adaptive algorithms can do. We have carefully tuned the hyper-parameters for each method and reported the best results in all the experiments.

\section{Background and Notations}
\label{sec: background}
\textbf{Overview of Adaptive Methods.} To compare AdaX with other optimization methods, we follow \citet{Reddi2018On} to present a generic framework of adaptive algorithms as shown in {Algorithm \ref{Algorithm_1}}.

Let $\mathcal{S}_{d}^+$ be the set of all positive symmetric definite matrices in $\mathbb{R}^{d\times d}$, and $\mathcal{F}$ be the parameter domain. For any adaptive algorithm, we first initialize parameters to be at $x_0$ and input the sequence of step sizes $\{\alpha_t\}_{t=1}^T$. In line 2 of {Algorithm \ref{Algorithm_1}},  $\phi_t: \mathcal{F} \rightarrow \mathbb{R}^d$ and $\psi_t:  \mathcal{F} \rightarrow \mathcal{S}^d_{+}$ are unspecified moment functions that calculate first and second moments. 
After obtaining the gradient at time $t$ in line 5, we can calculate the corresponding first and second moment $m_t$ and $V_t$ using these functions, where $m_t \in \mathbb{R}^{d}, V_t \in \mathcal{S}^d_+$. We then subtract $x_t$ by the update step $\alpha_t m_t/\sqrt{V_t}$.  Here, we use the element-wise square-root operation on $V_t$ and the division $m_t/\sqrt{V_t}$ is defined as $\sqrt{V_t}^{-1}m_t$.
The projection operation $\Pi_{\mathcal{F}, M}(y)$ in line 7 is defined as $\text{argmin}_{x \in \mathcal{F}}\|\sqrt{M}(x-y)\
\|$, where $M \in \mathcal{S}_{d}^+$ and $ y \in \mathbb{R}^d$, and it projects the updated parameters $x_t - \alpha_t m_t/\sqrt{{V}_t}$ back to the original parameter domain.

The main differences between the adaptive methods and the conventional SGD are in line 6 and 7. Specifically, SGD uses $\alpha_t g_t$ as the update step while in adaptive methods, the matrix $V_t$ scales the overall step size $\alpha_t$ element-wisely by $1/\sqrt{V_t}$, known as the adaptive learning rate. If we let $V_t$ be the identity matrix $\mathbb{I}$ and $\phi_t = g_t$, then algorithm \ref{Algorithm_1} becomes the vanilla SGD

\textbf{Adam and its Variants.} Using the general framework in Algorithm \ref{Algorithm_1}, we are able to summarize many adaptive optimization algorithms proposed recently. In most adaptive algorithms, $\phi_t$ is defined as
\begin{equation}
\begin{aligned}
    \phi_t = \beta_1 m_{t-1} + (1-\beta_1)g_t,
    \label{phit}
\end{aligned}
\end{equation}
where $\beta_1$ is the first moment parameter. In AdaGrad \citep{Duchi2011Adaptive} and RMSProp \citep{Tieleman2012RMSProp} , $\beta_1 = 0$ and $\phi_t = g_t$. When $\beta_1 \neq 0$, $\phi_t$ is similar to the momentum design of SGD \citep{Polyak1964Some} and many popular algorithms such as Adam \citep{Kingma2015Adam} and AMSGrad \citep{Reddi2018On} set $\beta_1 = 0.9$ in practice. 

On the other hand, different methods can have very different $\psi_t$'s. We provide a summary of different designs of existing  $\psi_t$'s in \textbf{Table \ref{Tab: Summary}}.
For instance, \citet{Duchi2011Adaptive} designed the $\psi_t$ in AdaGrad as the global average of  past gradient squares. However, recent algorithms such as RMSProp \citep{Tieleman2012RMSProp} and Adam \citep{Kingma2015Adam} chose the exponential moving average design instead. For Adam as an example, we have
\begin{equation} \tag{Adam}
    \begin{aligned}
  \psi_t = (\frac{1-\beta_2}{1-\beta_2^t})\text{diag}(\sum_{i=1}^t  \beta_2^{t-i} g_i^2),
   \end{aligned}
\end{equation}
where $\beta_2$ is the second moment parameter and $g_i^2$ denotes the element-wise square of the gradients. The diagonal operation $\text{diag()}$ performs the dimension transformation from $\mathbb{R}^d$ to $\mathcal{S}^d_{+}$.
To improve the generalization ability of Adam, other algorithms that slightly modify the second moment of Adam have been proposed. For example, \citet{Reddi2018On} proposed AMSGrad to take max operation on the second moment of Adam. \citet{Zhou2019AdaShift} argued that the $g_t^2$ in Adam's $V_t$ can be replaced with some past gradient squares $g_{t-n}^2$ to temporarily remove the correlation between the first and second moment and proposed AdaShift. \citet{Huang2019Nostalgic} changed the constant $\beta_2$ in Adam to a sequence of $\beta_{2(t)}$'s and constructed NosAdam. It was noticeable that these algorithms, due to their exponential moving average design, still assigned relatively high weights to recent gradients and past information was not emphasized. 
Besides, \citet{chen2018closing} noticed that replacing the square root operation in Adam by $p^{th}$($p < 1/2$) power could improve its performance and proposed Padam. However, they didn't change the exponential moving average design either.

\begin{algorithm}[t]
   \caption{Generic Adaptive Optimization Algorithm 
   }
   
   \label{Algorithm_1}
\begin{algorithmic}[1]
   \STATE \textbf{Input:} $x_0 \in \mathcal{F}$, sequence of step sizes $\{\alpha_t\}_{t=1}^T$
   \STATE \textbf{Moment functions}: $\{\phi_t, \psi_t\}_{t=1}^{T}$
   \STATE \textbf{Initialize} $m_0 = 0, V_0 = 0$
   \FOR{$ t= 1 $ \textbf{to} $T$}
   \STATE $g_t = \nabla f_t(x_t)$
   \STATE $m_t =  \phi_t(g_1, g_2, \dots, g_t), V_t =  \psi_t(g_1, g_2, \dots, g_t)$
   \STATE $ x_{t+1} = \Pi_{\mathcal{F}, \sqrt{V_t}}(x_t - \alpha_t m_t/\sqrt{{V}_t})$
   \ENDFOR
\end{algorithmic}
\end{algorithm}

\begin{table*}[t]
\caption{Comparisons of different designs of the second moment} 
\label{Tab: Summary}
\small
\centering
\begin{tabular}{ c| c c c}
 \hline
   & SGDM & AdaGrad & RMSProp \\ 
 \hline
 \\[-0.5em]
 $\psi_t$ & $\mathbb{I}$ & $\text{diag}(\sum_{i=1}^t g_i^2/t)$ & $\text{diag}(\sum_{i=1}^t  (1-\beta_2)\beta_2^{t-i} g_i^2)$
 \\[+1em]
  \hline
     & Adam & AMSGrad & AdaShift \\ 
 \hline
 \\[-0.5em]
 $\psi_t$ & $(\frac{1-\beta_2}{1-\beta_2^t})\text{diag}(\sum_{i=1}^t  \beta_2^{t-i} g_i^2)$ & $\text{diag}(\text{max}_t(\sum_{i=1}^t  (1-\beta_2)\beta_2^{t-i} g_i^2))$ & $\text{diag}(\sum_{i=1}^t  (1-\beta_2)\beta_2^{t-i} g_{i-n}^2)$
 \\[+1em]
   \hline
     & NosAdam & ...&\textbf{AdaX (ours)} \\ 
    \hline
\\[-0.5em]
 $\psi_t$ & \text{diag}$(\sum_{i=1}^t  \beta_{2i}\Pi_{j=1}^{t-i}(1+\beta_{2(t-j+1)}) g_i^2)$ & ... & $\frac{1}{(1+\beta_2)^t - 1}\text{diag}(\sum_{i=1}^t  \beta_2 (1+\beta_2)^{t-i} g_i^2)$\\
 \hline
\end{tabular}
\end{table*}

\textbf{Convex Convergence Analysis.} A commonly used framework for analyzing convex optimization algorithms was constructed by \citet{Zinkevich2003Online}.
In this framework, the optimization algorithm chooses a parameter set $\theta_t \in \mathcal{F}$ and an unknown convex cost function $f_t(\theta)$ evaluates its performance at $\theta_t$ in each iteration. Suppose that there exists a best parameter $\theta^*$ such that $\theta^* = \text{argmin}_{\theta \in \mathcal{F}}\left(\sum_{t=1}^T f_t(\theta)\right)$. Then a metric used to show the algorithm's performance is the regret function $R_T = \sum_{t=1}^T f_t(\theta_t) - f_t(\theta^*) $. The regret function accounts for the optimization speed since smaller $f_t(\theta_t) - f_t(\theta^*)$ represents $\theta_t$ is closer to the optimum $\theta^*$, and we want to ensure that $R_T/T \rightarrow 0$ so that the algorithm will always converge to the optimal solution. 

\textbf{Nonconvex Convergence Analysis.} There are many results for the convergence analysis of adaptive algorithms in the non-convex setting, such as \citet{Chen2019On} and \citet{zhou2018on}. 
We follow \citet{Chen2019On} to derive the convergence rate in this paper. Suppose we use an algorithm to minimize a cost function $f$ that satisfies the three assumptions below.

\begin{quote}
 \textbf{A1}. $f$ is differentiable and has $L$-Lipschitz gradient, i.e. $\forall x, y, \|\nabla f(x)- \nabla f(y)\| \leq L\|x - y\|$. and $f(x^*) > - \infty$, where $x^*$ is an optimal solution.  

\textbf{A2}. At time t, the algorithm obtains a bounded noisy gradient $g_t$ and the true gradient $\nabla f(x_t)$ is also bounded, i.e. $\|\nabla f(x_t)\| \leq G_{\infty}, \|g_t\| \leq G_{\infty}, \forall t \geq 1$ for some $G_\infty > 0$. Also, $\|\alpha_t\frac{ m_t}{\sqrt{V_t}}\| \leq G $ for some $G > 0$.

 \textbf{A3}. The noisy gradient is unbiased and the noise is independent, i.e. $g_t = \nabla f(x_t) + \eta_t, \mathbb{E}[\eta_t] = 0$ and $\eta_i$ is independent of $\eta_j$ if $i \neq j$.
\end{quote}

Then we ensure the convergence of the algorithm by showing that the norm of gradients approaches zero
\begin{equation}
    \begin{aligned}
   \min_{t\in [T]}\mathbb{E}\left[ \|\nabla f(x_t)\|^2 \right] = O(\frac{s_1(T)}{s_2(T)})
   \end{aligned}
\end{equation}
where $s_1(T), s_2(T)$ are functions of $T$ and $s_1(T) = o(s_2(T))$, which means $s_1(T)/s_2(T) \rightarrow 0$.

\textbf{Non-convergence of Adam in a Convex Setting.} \citet{Reddi2018On} proposed that the matrix $\Gamma_t$ defined as follows, was mistakenly assumed to be positive semi-definite in the original convergence proof of Adam.  

\begin{equation}
    \begin{aligned}
  \Gamma_t = \left( \frac{\sqrt{V_{t+1}}}{\alpha_{t+1}} - \frac{\sqrt{V_t}}{\alpha_t}\right),
   \end{aligned}
\end{equation}
where $V_t$ and $\alpha_t$ were the second momentum and step size at time $t$ as defined in \textbf{Algorithm \ref{Algorithm_1}}. Adam's regret function $R_T$ could be unbounded if the positive semi-definiteness was not satisfied. Based on such an observation, they constructed the following online convex optimization problem, in which Adam failed to converge to the optimal point. Let $C > 2$ be a fixed constant and $ \{f_t\}$ be the sequence of cost functions whose sum is to be minimized. Let
\begin{equation}
\label{non_convergence_problem}
f_t(x) = 
\begin{cases} 
Cx, ~~~\text{for } t \text{ mod } 3 = 1,\\
    -x, ~~~\text{otherwise},
  \end{cases}
\end{equation}
 where $x\in \mathcal{F} = [-1,1]$. It can be observed that the minimum regret is obtained at $x = -1$. However, correct large gradients (C) only appear every three iterations while noisy wrong gradients (-1) exist and can lead the algorithm away from the optimum. In this problem, SGD can counteract the effects of the noisy gradients and converge to the optimal solution. However, Adam can not distinguish between the correct gradient directions ($C$) and the noisy gradient directions ($-1$), because its $\sqrt{V_t}$ scales these gradients to be of similar sizes, which forces the algorithm to reach a highly suboptimal solution $x=1$ every three iterations. Such a problem reveals the fact that Adam's design of adaptive learning rate is problematic, but it is also questionable whether such a high level of noise exists in real situations.

\section{Problem of Adam Revisited}
\label{sec: problem}
In this section, we further discuss the problem of Adam by changing the non-convergence problem (\ref{non_convergence_problem}) to a more practical setting and explain why the fast convergence of Adam impairs its performance in the long term. We use the following synthetic example to show why we need to completely change the exponential moving average design. Consider the simple sequence of convex continuous functions $\{f_t\}$
\begin{equation}
\begin{aligned}
\label{non_convergence_problem_2}
&f_t(x) = |C \lambda^{t-1}x|  &\forall t \leq n. \\
&f_t(x) =
    \begin{cases}
     C \lambda^{t-1}x, ~~~\text{ for } x \geq 0,\\
     0, ~~~\text{ for } x < 0 ,
   \end{cases}
   &\forall t > n.
\end{aligned}
\end{equation}
where $C \in (1,+\infty), \lambda \in (0,1) $ are constants in $\mathbb{R}$. $n$ is a small integer to make $x=0$ the only optimum in this problem and a simple choice is $n=1$. We set the parameter domain $\mathcal{F}= [-2, C/(1-\lambda)]$. Suppose we initialize at some $x_0 > 0$, then this problem simulates a situation where the gradient decreases exponentially as time increases, implying that the algorithm is approaching the global minimum where smaller updates are needed.  Such a phenomenon can also be observed in real training processes, which will be shown in the experiments. 

Unlike the problem in Eqn.(\ref{non_convergence_problem}), no high frequency noise exists in the gradients of problem (\ref{non_convergence_problem_2}). However,no gradients exist at $x<0$ when $t>n$ and thus no algorithm could escape if they enter this region. We are interested in whether different algorithms has the ability to converge to 0, if the initial step size $\alpha_0$ and the initialization $x_0 > 0 $ can be arbitrarily chosen. Since the gradients follow a geometric sequence, the sum is always finite and hence no matter where $x_0$ is initialized, there always exists an $\alpha_t$ such that SGD converges to 0. However, Adam does not have a similar property. We propose the following theorem.

\begin{theorem} In problem (\ref{non_convergence_problem_2}), with $\beta_1 = 0, \beta_2 \in(0, \lambda^2)$ in Adam and $\alpha_t \geq {\alpha}/{t}, \forall x_0 > 0, \forall \alpha_1 > 0, $ Adam can never converge to 0, i.e. $\exists T > 0, \text{ s.t. } x_T < 0$, 
\label{Thm: Non_convergence}
\end{theorem}

We relegate all the proofs to the Appendix. 
In the analysis, the condition $\beta_1 < \sqrt{\beta_2}$ for proving the convergence of Adam mentioned by \citet{Kingma2015Adam} is automatically satisfied.
Besides, $\alpha_t \geq \alpha/t$ is a weak requirement for the step sizes and it can be ensured with constant step sizes or $\alpha_t = \alpha/\sqrt{t}$ as in the convergence analysis in section \ref{sec: our_approach}.

The theorem claims that however close Adam is initialized to zero and however small a initial learning rate is chosen, Adam always goes to the negative region. Intuitively, Adam scales the exponentially decreasing gradient by $1/\sqrt{V_t}$ at each iteration. However, due to its exponential moving average design, the second moment $V_t$ only focuses on recent small gradient squares and is also very small at the same time. Therefore, Adam's $g_t/\sqrt{V_t}$ would be larger than a fixed constant at any time step and would ultimately lead the algorithm to the negative region regardless of initialization. One may wonder whether the first moment design helps Adam in such a situation. However, we also show that as long as the condition $\beta_1 < \sqrt{\beta_2}$ is satisfied, Adam would always goes to $x<0$ for any $\beta_1 > 0$. Therefore, although Adam converges faster than SGD due to its large updates, it cannot slow down when approaching the optimum.

\textbf{Problem of Adam Variants.} We choose AMSGrad as an example to show why current variants of Adam do not solve Adam's problem completely. As mentioned by \citet{Reddi2018On}, AMSGrad was constructed to address the problem of Adam's large steps, by replacing the second moment $V_t$ by its maximum in time, denoted as $\hat{V}_t$. Such a design may be useful in certain cases since it keeps some of the past memory and prevents $V_t$ from being too small. However, the time for achieving $\text{max}(\hat{V}_t)$ is task-dependent. The following theorem proves that for certain cases in problem (\ref{non_convergence_problem_2}), AMSGrad is incapable of improving Adam.
\begin{theorem} In problem (\ref{non_convergence_problem_2}) with $\alpha_t \geq {\alpha}/{t}$, $\forall \beta_2 \in (0,1), \exists \lambda \in (\sqrt{\beta_2}, 1)$, such that  AMSGrad can never converge to 0, i.e.  $\forall x_0 > 0, \forall \alpha_1 > 0, \exists T > 0, \text{ s.t. } x_T < 0$.
\label{Thm: Non_convergence_AMS}
\end{theorem}
The above theorem essentially states for any fixed $\beta_2$, we can find a $\lambda$ such that AMSGrad cannot help Adam. The reason is that its $V_t$ keeps increasing before stepping into the negative region, which makes the maximum operation ineffective and AMSGrad performs exactly the same as Adam. Therefore, the effectiveness of AMSGrad depends on the situation and the source of the problem, that is, the exponential moving average has not completely changed. Other Adam variants, such as AdaShift \citep{Zhou2019AdaShift} have a similar issue. We also show that such fast-decreasing gradients exist in real applications in the experiments. Therefore, to completely get rid of the problem of Adam, we need a totally different design of the adaptive learning rate and control the update steps effectively. The above analysis provides some intuition on why Adam variants trains much faster than SGD, but cannot have comparable testing performance.

\section{Our Approach}
\label{sec: our_approach}
We introduce our novel optimization algorithm to adjust the adaptive learning rate.  Based upon the above discussions, we know that small gradients can generate unstable second moment and that past memory should be emphasized (such as the max operation in AMSGrad). Moreover, the emphasis operation should not be task-dependent. To counteract the exponential decrease in gradients, we propose to do exactly the opposite of Adam by weighting exponentially more on the past gradients and gradually decrease the adaptivity to current gradients, as shown in \textbf{Algorithm \ref{Algorithm_2}}. The most important differences between AdaX and Adam are in line 6 and 7, where instead of using an exponential moving average, we change $(\beta_2, 1-\beta_2)$ to $(1+\beta_2, \beta_2)$ in our design. In line 6, we can see that past gradients are multiplied by a constant larger than $1$, which means that past information is accumulated rather than forgotten. Each $g_t^2$ is still multiplied by a small number and added to the past memory. The intuition behind our algorithm is that we want to gradually decrease the adaptivity of the second moment to the latest gradients because they become sparse and noisy when the parameters are close to the optimal points, which is similar to the synthetic example (\ref{non_convergence_problem_2}). Our design guarantees such small gradients cannot greatly influence the update steps when a large $v_t$ is maintained. With the bias correction term, our $\hat{v}_t$ will gradually become stable and large.

\begin{algorithm}[t]
   \caption{AdaX Algorithm}
   \label{Algorithm_2}
\begin{algorithmic}[1]
   \STATE \textbf{Input:} $x \in \mathcal{F}$, $\{\alpha_t\}_{t=1}^{T}, (\beta_{1}, \beta_{2}) = (0.9, 10^{-4})$
   \STATE \textbf{Initialize} $m_0 = 0, v_0 = 0$
   \FOR{$ t= 1 $ \textbf{to} $T$}
   \STATE $g_t = \nabla f_t(x_t)$
   \STATE $m_t =  \beta_{1}m_{t-1} + (1-\beta_1) g_t$
   \STATE $v_t =  (1+\beta_{2})v_{t-1} + \beta_{2} g_t^2$
   \STATE $\hat{v}_t = v_t / [(1+\beta_{2})^t -1]$ and $V_t = \text{diag}(\hat{v}_t)$
   \STATE  $x_{t+1} = \Pi_{\mathcal{F}, \sqrt{V_t}}(x_t - \alpha_t m_t/\sqrt{\hat{v}_t})$
   \ENDFOR
\end{algorithmic}
\end{algorithm}

In line 7, in order to achieve an unbiased estimate of second moment, we divide our $v_t$ by the bias correction term. Similar to \citet{Kingma2015Adam}'s derivation, let $g_t$ be the gradient at time step $t$ and further suppose $g_t$'s are drawn from a stationary distribution $g_t \sim p(g_t)$. By taking expectation on both sides of line 6 in Algorithm \ref{Algorithm_2}, we get
\begin{equation}
    \begin{aligned}
   \mathbb{E}({v_t})= \sum_{i=1}^t (1+\beta_2)^{t-i} \beta_2 \mathbb{E}(g_t^2) 
        = [(1+ \beta_2)^t - 1] \mathbb{E}(g_t^2). \nonumber
   \end{aligned}
\end{equation}

Therefore, to maintain an accurate second moment, we would divide $v_t$ by $(1+\beta_{2})^t -1$ in line 7. However, it's worth mentioning that we do not include a first moment correction term $(1-\beta_1^t)$ as in \citet{Kingma2015Adam} for the following reason. Consider the momentum in Stochastic Gradient Descent (SGDM) and Adam's first moment, 
\begin{equation}\nonumber
    \begin{aligned}
    \text{SGDM: }&m_t = \gamma m_{t-1} + g_t = \sum_{i=1}^t \gamma^{t-i} g_i, \\
    \text{Adam: }&m_t = \beta_1 m_{t-1} + (1-\beta_1)g_t = (1-\beta_1)\sum_{i=1}^t \beta_1^{t-i} g_i.
   \end{aligned}
\end{equation}
It can be observed that they have the same form except for the constant $1-\beta_1$, and therefore the first order bias correction term is counter-intuitive. Next, we show that our algorithm ensures the positive semi-definiteness of $\Gamma_t$ and hence does not have the non-convergence issue of Adam. 
\begin{lemma} Algorithm \ref{Algorithm_2} ensures that the matrix $\frac{V_{t}}{\alpha_t^2} - \frac{V_{t-1}}{\alpha_{t-1}^2} \succeq 0$
\label{Lem: PSD_of_V}
\end{lemma}
We also prove that AdaX can converge to the global minimum in our problem (\ref{non_convergence_problem_2}) in the Appendix section \ref{App: Proofs_of_nonconvergence_problem}. Finally, we provide the convergence analysis of our algorithm in both the convex and non-convex settings. Using the analysis framework by \citet{Zinkevich2003Online} in section \ref{sec: background}, we show that our AdaX algorithm has a regret bound $R_T = o(T)$.

\begin{theorem}
\label{Thm: Regret_Bound} Let $\{x_t\}$ and $\{v_t\}$ be the sequences obtained from Algorithm \ref{Algorithm_2}, $\alpha_t = \alpha/\sqrt{t}, \beta_{1,1} = \beta_1, \beta_{1,t} \leq \beta_1$, for all $t \in [T]$ and $\beta_{2t} = \beta_2/t$. Assume that $\mathcal{F}$ has bounded diameter $\|x-y\|_{\infty} \leq D_{\infty}, \forall x, y \in \mathcal{F}$ and $\|\nabla f_t(x)\| \leq G_{\infty} $ for all $t \in [T]$ and $x \in \mathcal{F}$.  Then for $x_t$ generated using Algorithm \ref{Algorithm_2}, we have the following bound on the regret.
\begin{equation}
    \begin{aligned}
   R_T & \leq \frac{D_{\infty}^2}{2\alpha_T(1-\beta_1)}\sum_{i=1}^d \hat{v}_{T,i}^{1/2} + \frac{D_{\infty}^2}{2(1-\beta_1)} \sum_{t=1}^T  \sum_{i=1}^d \frac{\beta_{1t}\hat{v}_{t,i}^{1/2}}{\alpha_t} \\
   &+  \frac{\alpha C \sqrt{T}}{(1-\beta_1)^3\sqrt{\beta_2 }} \sum_{i=1}^d \|g_{1:T, i}\|_2 \\
   \end{aligned}
\end{equation}
where $C$ is a constant that doesn't depend on $T$.
\end{theorem}
The following corollary follows naturally from the above theorem.

\begin{corollary}
\label{Cor: Regret_Bound} Suppose $\beta_{1t}  = \beta_1 \lambda^{t-1}$ in Theorem \ref{Thm: Regret_Bound}, then we have
\begin{equation}
    \begin{aligned}
   R_T & \leq \frac{D_{\infty}^2\sqrt{T}}{2\alpha(1-\beta_1)}\sum_{i=1}^d \hat{v}_{T,i}^{1/2} + \frac{d\beta_1D_{\infty}^2 G_{\infty}}{2\alpha(1-\beta_1)(1-\lambda)^2} \\
   & + \frac{\alpha C \sqrt{T}}{(1-\beta_1)^3\sqrt{\beta_2 }} \sum_{i=1}^d \|g_{1:T, i}\|_2 \\
   \end{aligned}
\end{equation}
\end{corollary}
The above theorem and corollary guarantee the convergence of AdaX when $\sum_{i=1}^d \hat{v}_{T,i}^{1/2} \ll \sqrt{d}$ and $ \sum_{i=1}^d \|g_{1:T, i}\|_2 \ll \sqrt{dT}$ \cite{Duchi2011Adaptive}. To compare the convergence speed in the non-convex setting, suppose we want to minimize a cost function $f$ satisfying the three assumptions \textbf{A1, A2, A3} in section \ref{sec: background}. Then we can obtain the following theorem, which proves that AdaX converges with a speed close to AMSGrad as mentioned by \citet{Chen2019On}.

\begin{theorem}
\label{Thm: Stationary} Let $\{x_t\}$ and $\{v_t\}$ be the sequences obtained from Algorithm \ref{Algorithm_2}, $\alpha_t = \alpha/\sqrt{t}, \beta_{1,1} = \beta_1, \beta_{1,t} \leq \beta_1$, for all $t \in [T]$ and $\beta_{2t} = \beta_2$. Assume that $\|\nabla f_t(x)\| \leq G_{\infty} $ for all $t \in [T]$ and $x \in \mathcal{F}$.  Then for $x_t$ generated using Algorithm \ref{Algorithm_2}, we have the following bound.
\begin{equation}
    \begin{aligned}
   \min_{t\in [T]}\mathbb{E}\left[ \|\nabla f(x_t)\|^2 \right] 
   &\leq \frac{G_{\infty}}{\alpha \sqrt{T}} (\frac{C_1G_{\infty}^2 \alpha^2}{c^2}(1+\log T) \\
  &+ \frac{C_2 d \alpha}{c} + \frac{C_3 d \alpha^2}{c^2} + C_4)
   \end{aligned}
\end{equation}
where $C_1, C_2, C_3, C_4$ are constants independent of $T$

\end{theorem}

Hence the convergence rate of AdaX is $O({\log T}/{\sqrt{T}})$.

\section{Experiments}
\label{sec: Experiments}
In this section, we present extensive experiments to examine the effectiveness and robustness of AdaX. Following \citet{Loshchilov2019Decoupled}, we use decoupled weight decay in all the adaptive algorithms. AdamW, AdaX-W refer to the Adam and AdaX algorithm with decoupled weight decay. We relegate the detailed implementation of AdaX-W to section \ref{App: implementation} in the Appendix \footnote{The code can be found in \href{https://github.com/switchablenorms/adax}{this repository}}.

\begin{figure*}[!ht]
\centering
\subfigure[\footnotesize Training Top-1 Accuracy on CIFAR-10]{
  \centering
  \includegraphics[width=0.23\linewidth]{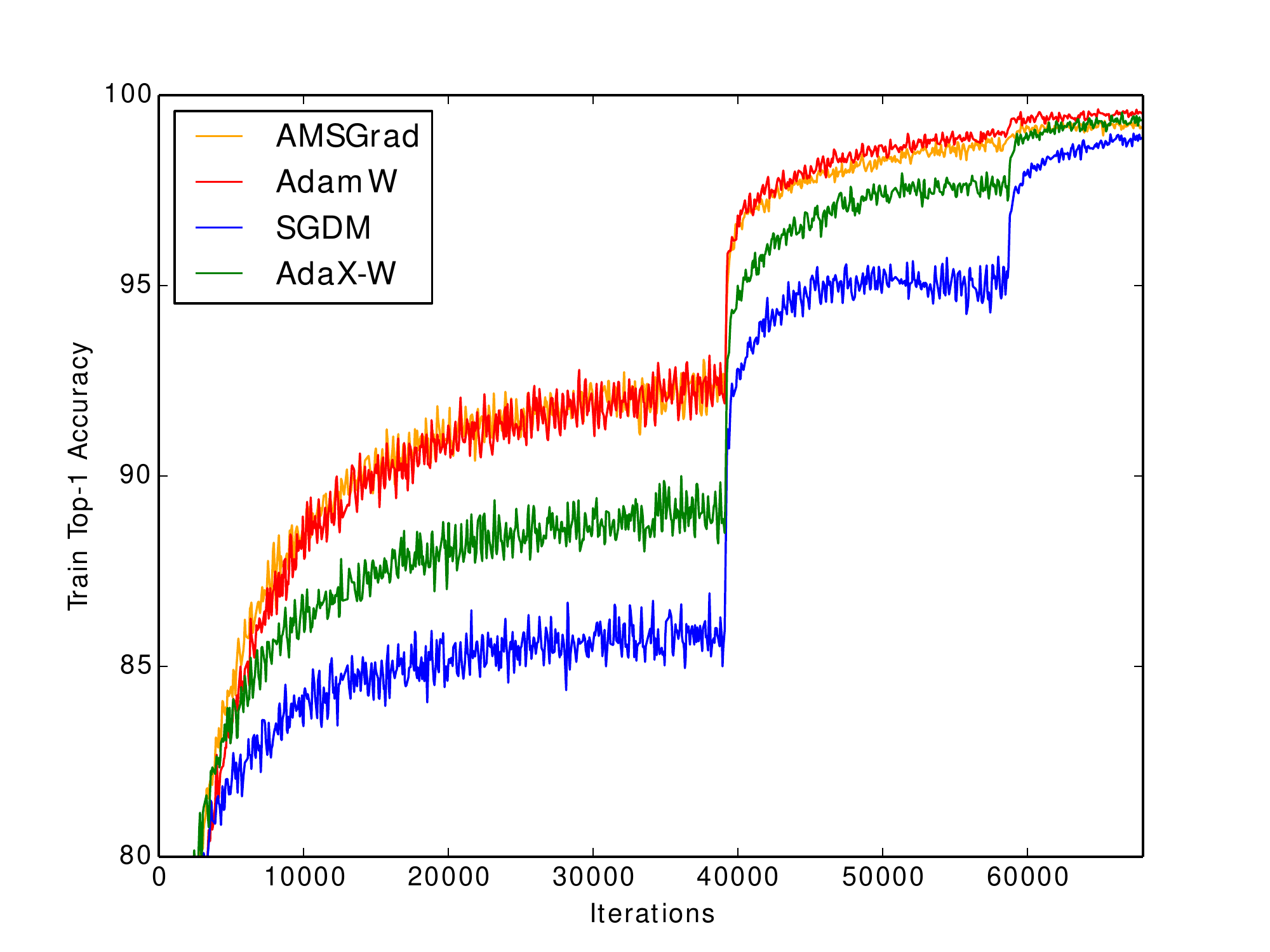}
  \label{fig: Cifar_1}
}
\subfigure[\footnotesize Testing Top-1 Accuracy on CIFAR-10]{
  \centering
  \includegraphics[width=0.23\linewidth]{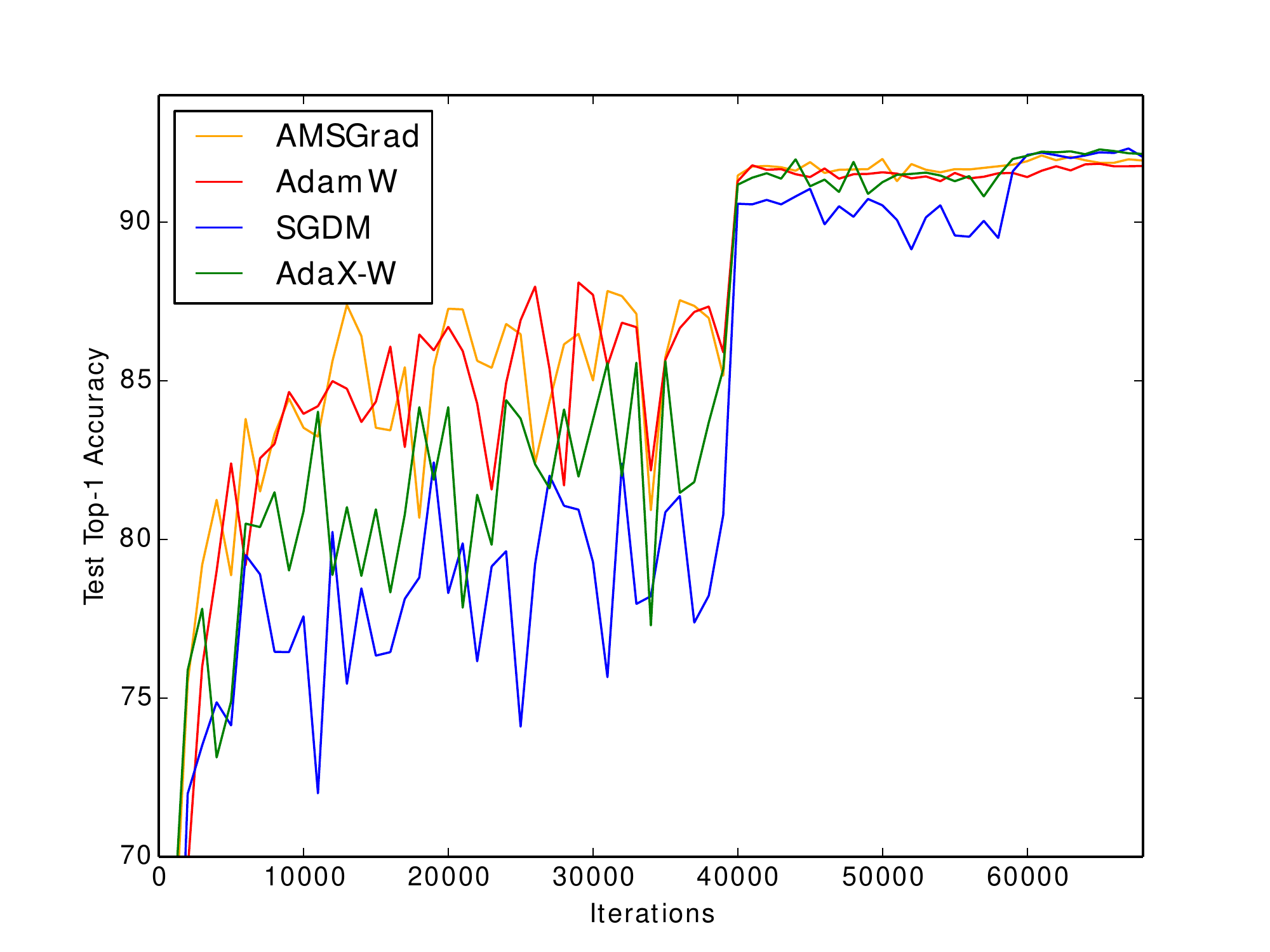}
  \label{fig: Cifar_2}
}
\subfigure[Training Top-1 Accuracy on ImageNet]{
  \centering
  \includegraphics[width=0.23\linewidth]{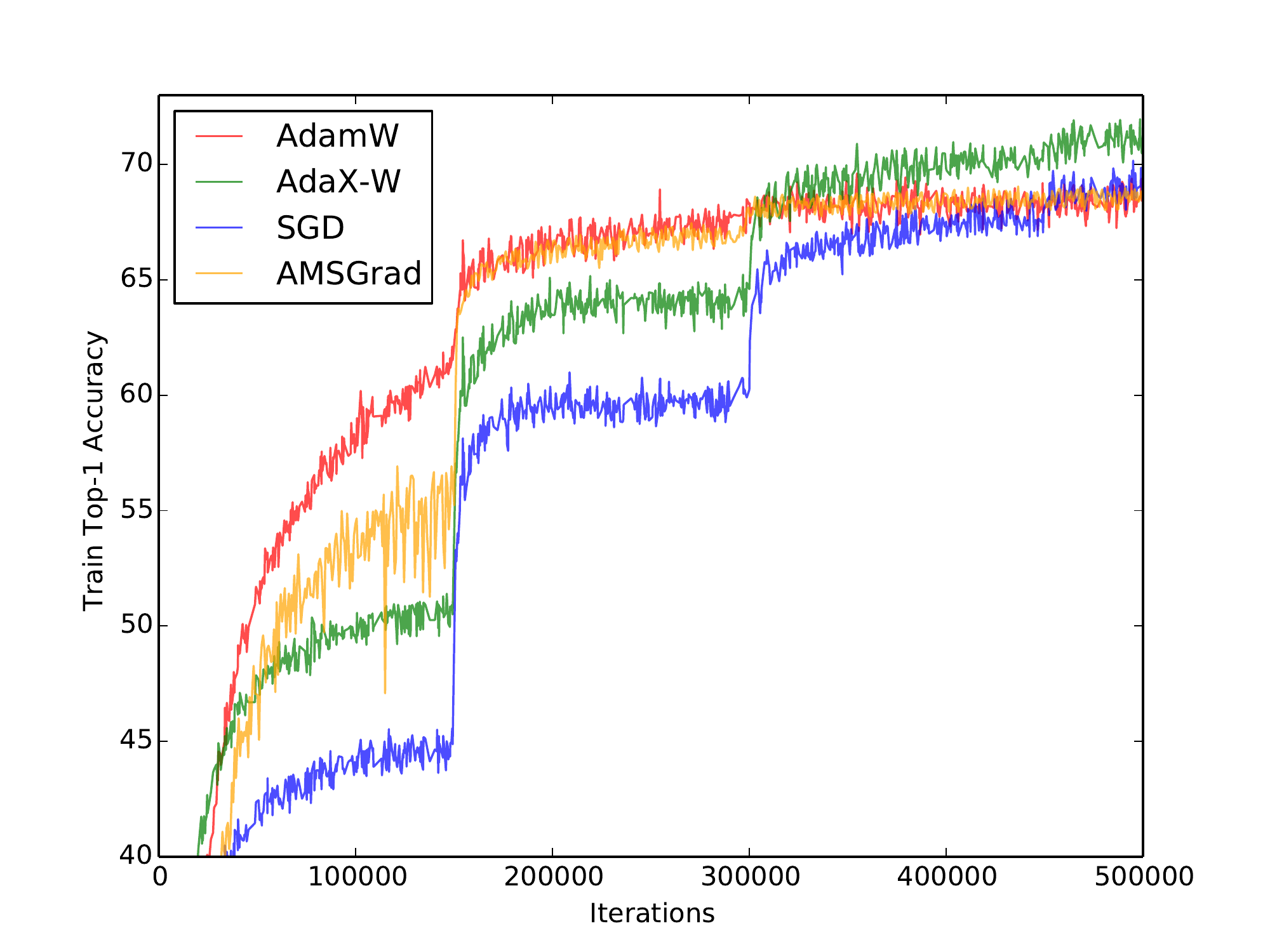}
  \label{fig: IN_train}
}
\subfigure[Testing Top-1 Accuracy on ImageNet]{
  \centering
  \includegraphics[width=0.23\linewidth]{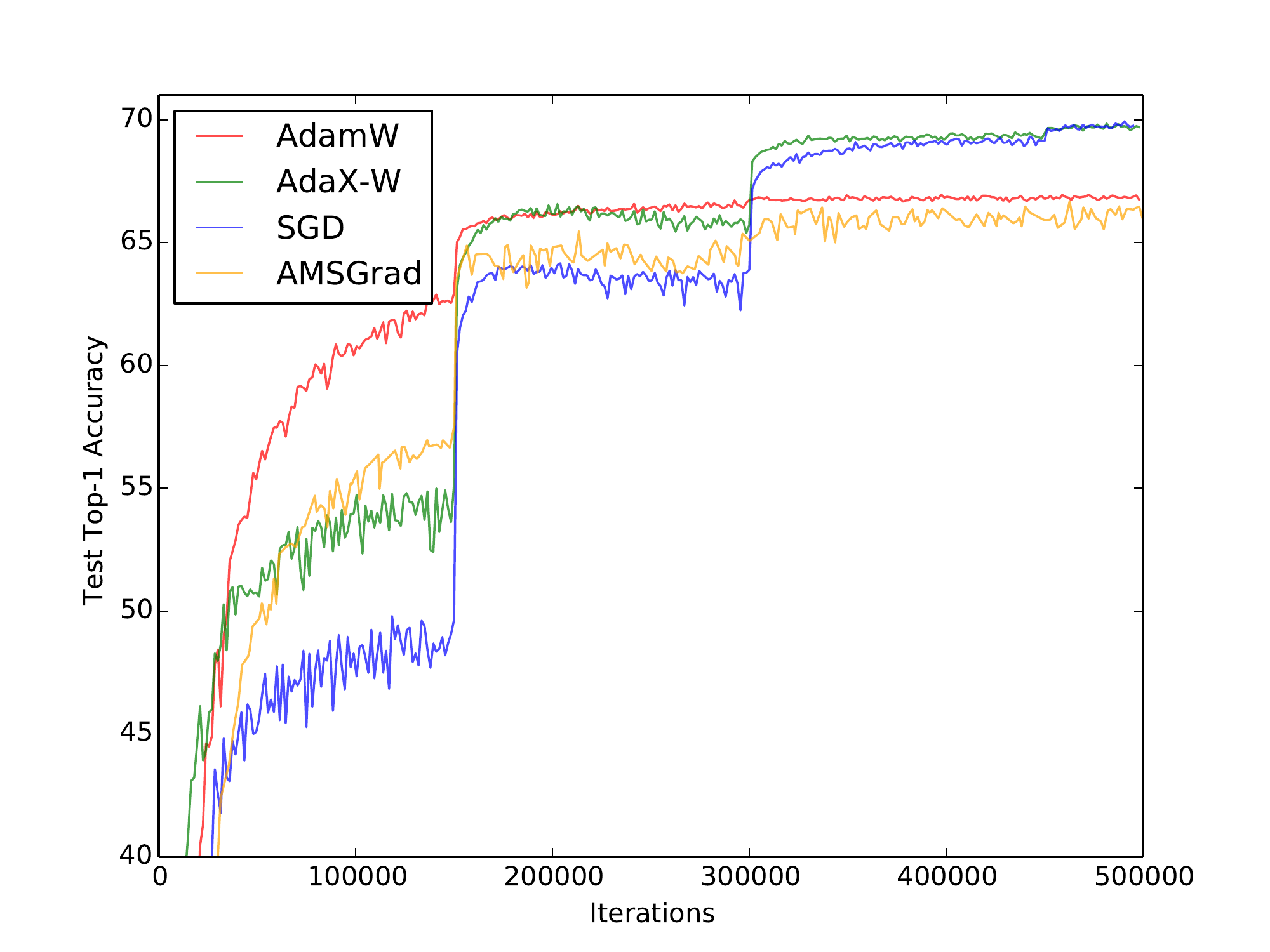}
   \label{fig: IN_test}
}
\label{fig: Cifar-10}
\caption{Training and Testing curves on CIFAR-10 and ImageNet}
\end{figure*}

\subsection{Performance Comparisons}
\label{subsec: performance_comparison}
We first evaluated the performance of AdaX-W in comparison with SGD with momentum (SGDM), AdamW \cite{Loshchilov2019Decoupled, Kingma2015Adam}, and AMSGrad \cite{Reddi2018On} on different deep learning tasks. As analyzed in section \ref{sec: problem}, Adam's unstable second moment led to its fast convergence in the experiments, but it also impaired its final performance and possibly trapped the algorithm in local minimums. The experiments below verified our claim that such instability was harmful to Adam's generalization performance and our new method could completely eradicate this instability. We thoroughly tuned the hyper-parameters in all the experiments and reported the best results for all the algorithms to ensure fair comparisons. The tuning details were provided in section \ref{App: Parameter_Tuning} in the Appendix. All our experiments were run on Nvidia-Tesla V100 GPUs.


\textbf{Image Classification on CIFAR-10}.
Using ResNet-20 created by \citet{He2016Deep}, we evaluated the performance of AdaX-W on the CIFAR-10 \cite{Cifar} dataset. A learning rate schedule that scaled down step sizes by $0.1$ at the $100$-th and the $150$-th epoch was utilized in training. As can be observed in Figure \ref{fig: Cifar_1}, \ref{fig: Cifar_2}, despite the initial super convergence of Adam and AMSGrad, their final accuracy could not catch up with the other two algorithms. On the other hand, AdaX-W converged faster than SGDM and yielded similar performances (92.32). We listed the average final accuracy over 3 independent runs in Table \ref{tab:IN_VOC}.


\textbf{Image Classification on ImageNet}.
We also conducted experiments to examine the performance of AdaX-W on ImageNet \cite{Deng2009Imagenet}. The famous ResNet-18 \cite{He2016Deep} model was used in training and a warm up scheme was applied in the initial 25k iterations \cite{Goyal2017Accurate}, and then the step size was multiplied by 0.1 at the 150k, 300k and 450k\textit{-th} iteration steps. As observed from Figure \ref{fig: IN_train} and \ref{fig: IN_test}, although AdamW was fast at the beginning, its test accuracy stagnated after the second learning rate decrease. AMSGrad performed even worse than AdamW. AdaX-W, on the other hand, converged faster than SGDM without loss of testing accuracy (69.87), as shown in Table \ref{tab:IN_VOC}. Moreover, AMSGrad had a much higher training accuracy than SGDM. 

\begin{table*}
\centering
\footnotesize
\caption{\footnotesize Validation Top-1 accuracy on CIFAR-10, ImageNet and IoU on VOC2012 Segmentation. The different algorithms were trained for the same number of iterations and the results were collected for 5 independent runs. We reported the mean accuracy and IoU as well as the margin of error. The two best results of each column were shown in bold.}
\begin{tabular}[t]{ c|c|c| c }
\hline
Dataset & CIFAR-10 & {ImageNet}  &VOC2012 Segmentation \\
 \hline
Model & ResNet-20 & ResNet-18  & Deeplab-ASPP \\
\hline
SGDM & \textbf{92.30 $\pm$ 0.09} &\textbf{69.90 $\pm$ 0.04} &\textbf{76.28 $\pm$ 0.15}\\
AdamW & {91.86 $\pm$ 0.04} &66.92 $\pm$ 0.05  & 74.62 $\pm$ 0.12 \\
AMSGrad(W) & 92.04 $\pm$ 0.05&66.64 $\pm$ 0.03 &73.62 $\pm$ 0.09\\
\textbf{AdaX-W(ours)}& \textbf{92.32 $\pm$ 0.04} & \textbf{69.87 $\pm$ 0.05}  & \textbf{76.53 $\pm$ 0.14}\\
\hline
\end{tabular}
\label{tab:IN_VOC}
\end{table*}

\begin{table}
\small
\flushright
\centering
\caption{\footnotesize  Validation perplexity on One Billion Word for language modeling. The results were collected for 3 independent runs and the best result was shown in bold. We reported the mean validation perplexity as well as the margin of error.}
\begin{tabular}[t]{ c| c }
\hline
 Method & Validation PPL \\
 \hline
AMSGrad &  61.66 $\pm$ 0.10 \\
Adam &  36.90 $\pm$ 0.05\\
\textbf{AdaX(ours)} &  \textbf{35.22 $\pm$ 0.07}\\ 
\hline
\end{tabular} 
\label{tab:lm}
\end{table}

\begin{figure*}
\centering
\subfigure[\footnotesize Training Dynamics on One Billion Word.]{
  \centering
  \includegraphics[width=0.3\linewidth]{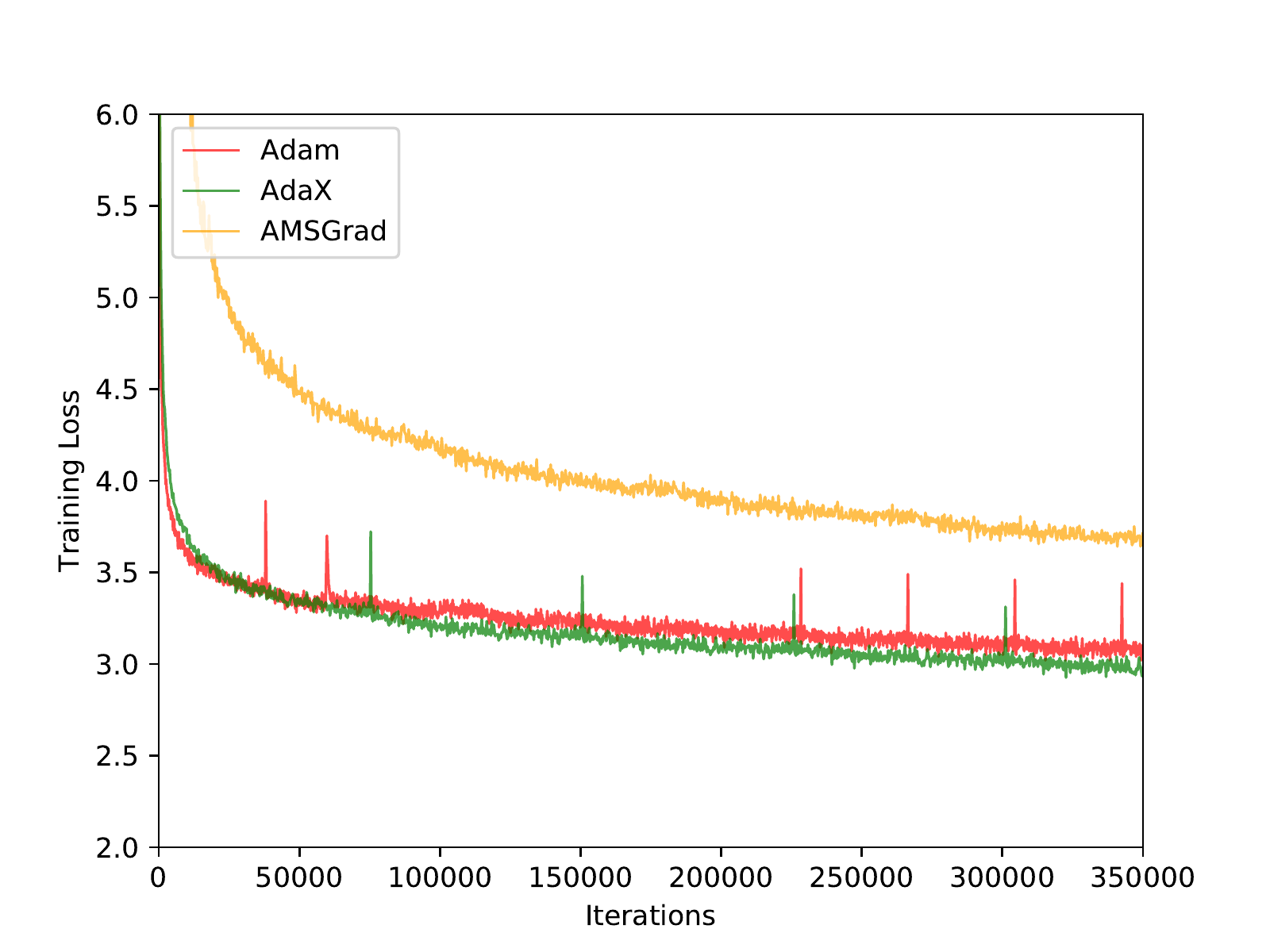}
  \label{fig: BillionWords}
}
\subfigure[\footnotesize Training Loss on VOC2012 Segmentation]{
  \centering
  \includegraphics[width=0.3\linewidth]{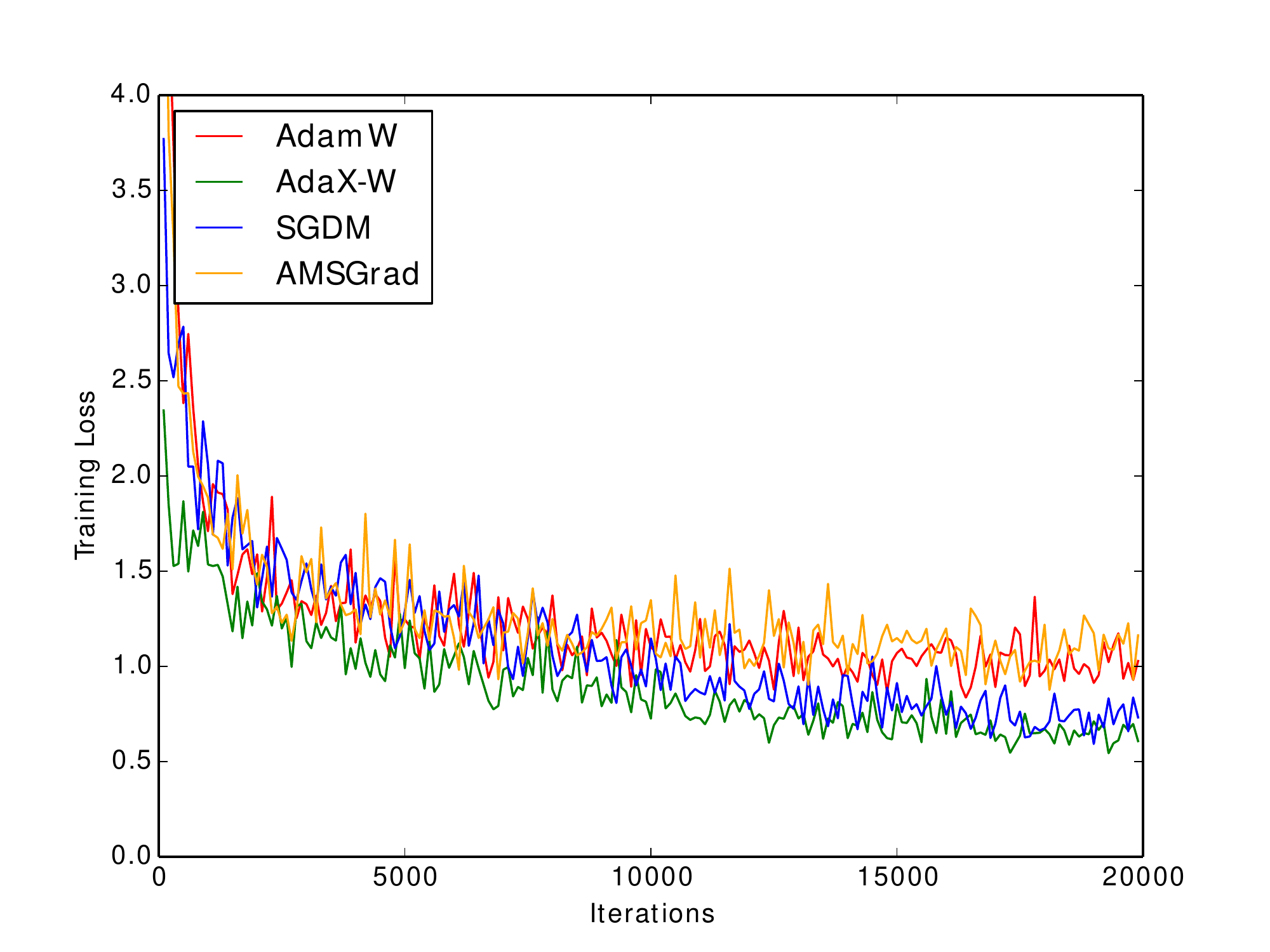}
  \label{fig: voc_train}
}
\subfigure[\footnotesize Testing IoU on VOC2012 Segmentation]{
  \centering
  \includegraphics[width=0.3\linewidth]{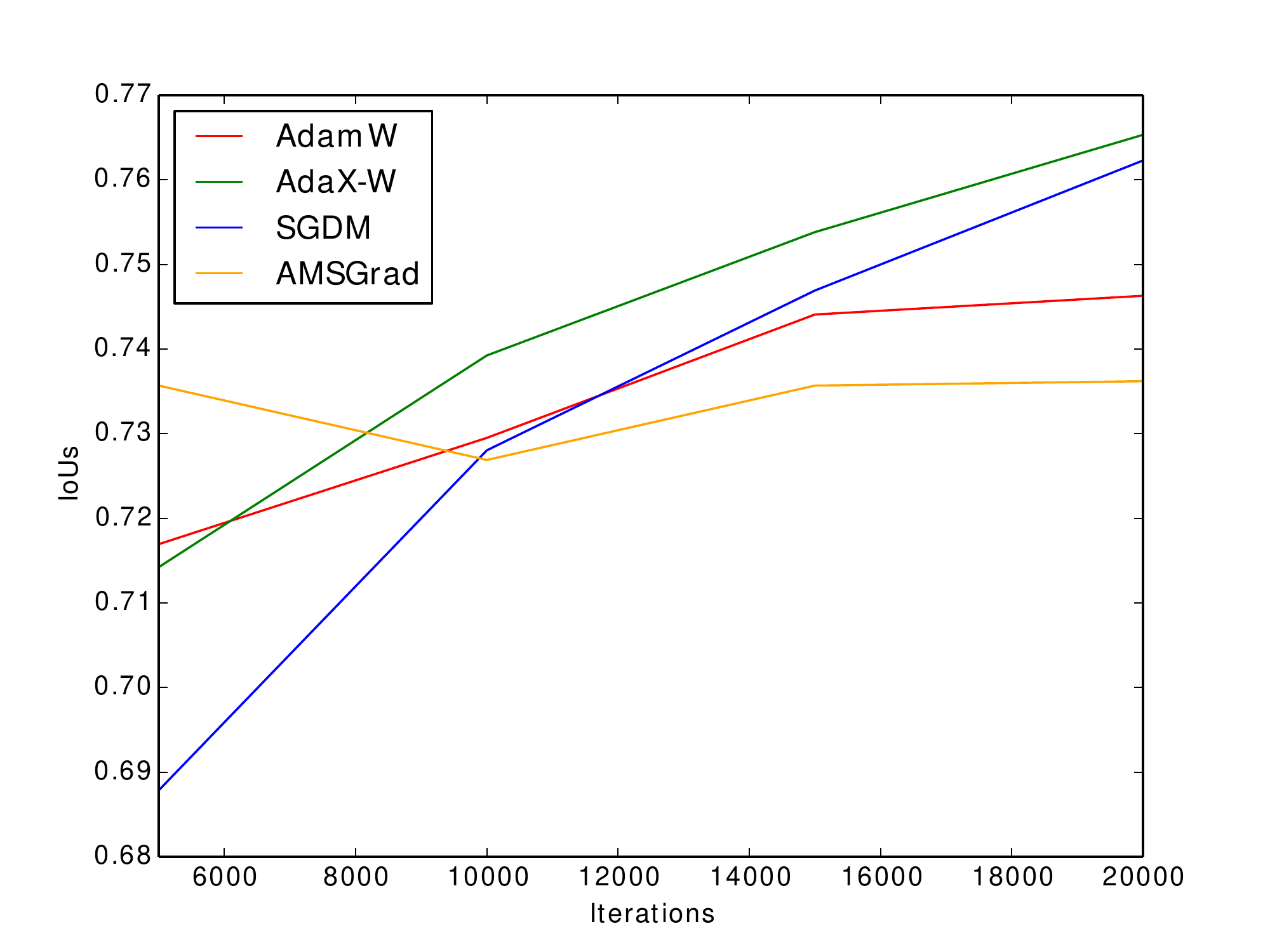}
  \label{fig: voc_test}
}
\caption{(a)~ Traning loss curves for different algorithms on One Billion Word. (b, c)~ Training Loss and Testing Intersection over Union (IoU) on the VOC2012 Segmentation task.}
\end{figure*}

\textbf{Language Modeling.}
AdaX was also validated on One Billion Word \cite{ chelba2013one} dataset of language modeling task.  For the One Billion Word, we used a two-layer LSTMs with 2048 hidden states and sampled softmax. The experiment settings in the publicly released code  \citet{rdspring1} was adopted in this study. For vanilla Adam, AMSGrad and AdaX, the LSTMs were trained for 5 epochs, with learning rate decaying to 1e-8 linearly. Note that SGDM was not suitable in this task due to the presence of sparse gradients \citep{Duchi2011Adaptive}, so it was not included in the comparison. 
The training loss and the validation perplexity were shown in Figure \ref{fig: BillionWords} and Table \ref{tab:lm}. We could see that the AdaX outperformed the Adam baseline by a significant margin (35.22  vs. 36.90). Moreover, AdaX started a little slower at the early stage, but it soon surpassed Adam on both training and validation performance, which corresponded to our claim that Adam's super convergence was harmful. AMSGrad, on the other hand, performed rather poorly in this task.

\textbf{Transfer Learning}.
To further examine the effectiveness of AdaX in transfer learnings such as semantic segmentation, we evaluated its performance on the PASCAL VOC2012 augmented dataset \cite{VOC2012} \cite{VOC2012Aug}. The Deeplab-ASPP model proposed by \citet{Chen2016Deeplab} with a ResNet-101 backbone pretrained on the MS-COCO dataset\cite{MSCOCO} was adopted. 
We evaluated the algorithms' performances at the 5k, 10k, 15k and 20k iterations using intersection over union (IoU). As could be observed in Figure \ref{fig: voc_test} and Table \ref{tab:IN_VOC}, AdaX-W trained faster than SGDM and obtained a higher IoU (76.5) at the same time. However, AdamW and AMSGrad could not obtain comparable results.

\subsection{Stability of Second moment Design}
Besides, we compared the second moment design of Adam and AdaX empirically and proved the existence and influence of the instability of Adam's second moment. We also showed that our design was stable and robust. 

\begin{figure*}[ht]
\centering
\subfigure[\footnotesize Updates over Iterations]{
  \centering
  \includegraphics[width=0.23\linewidth]{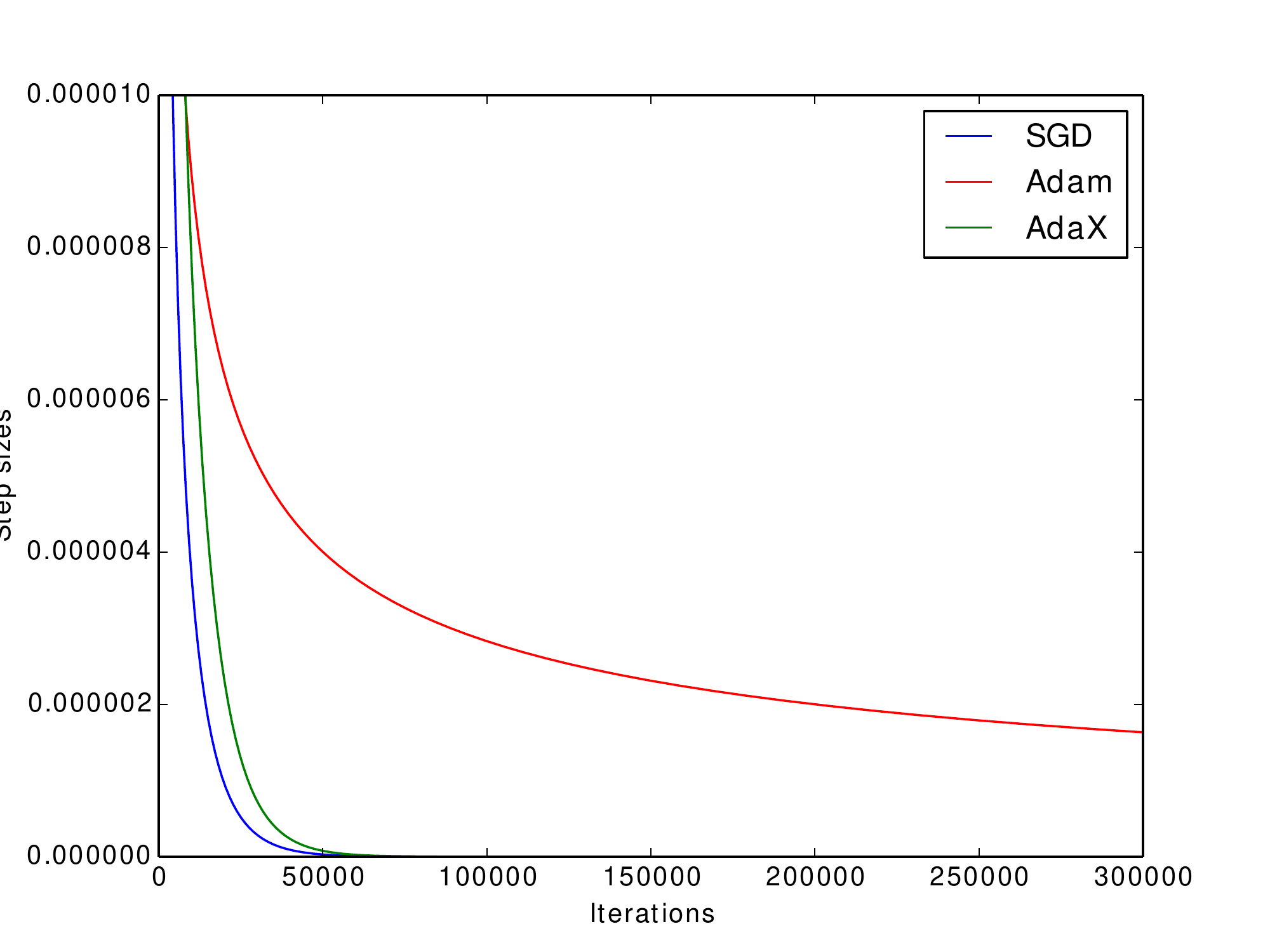}
  \label{fig: toy_update}
}
\subfigure[\footnotesize Average Second Moment over Iterations on CIFAR-10]{
  \centering
  \includegraphics[width=0.23\linewidth]{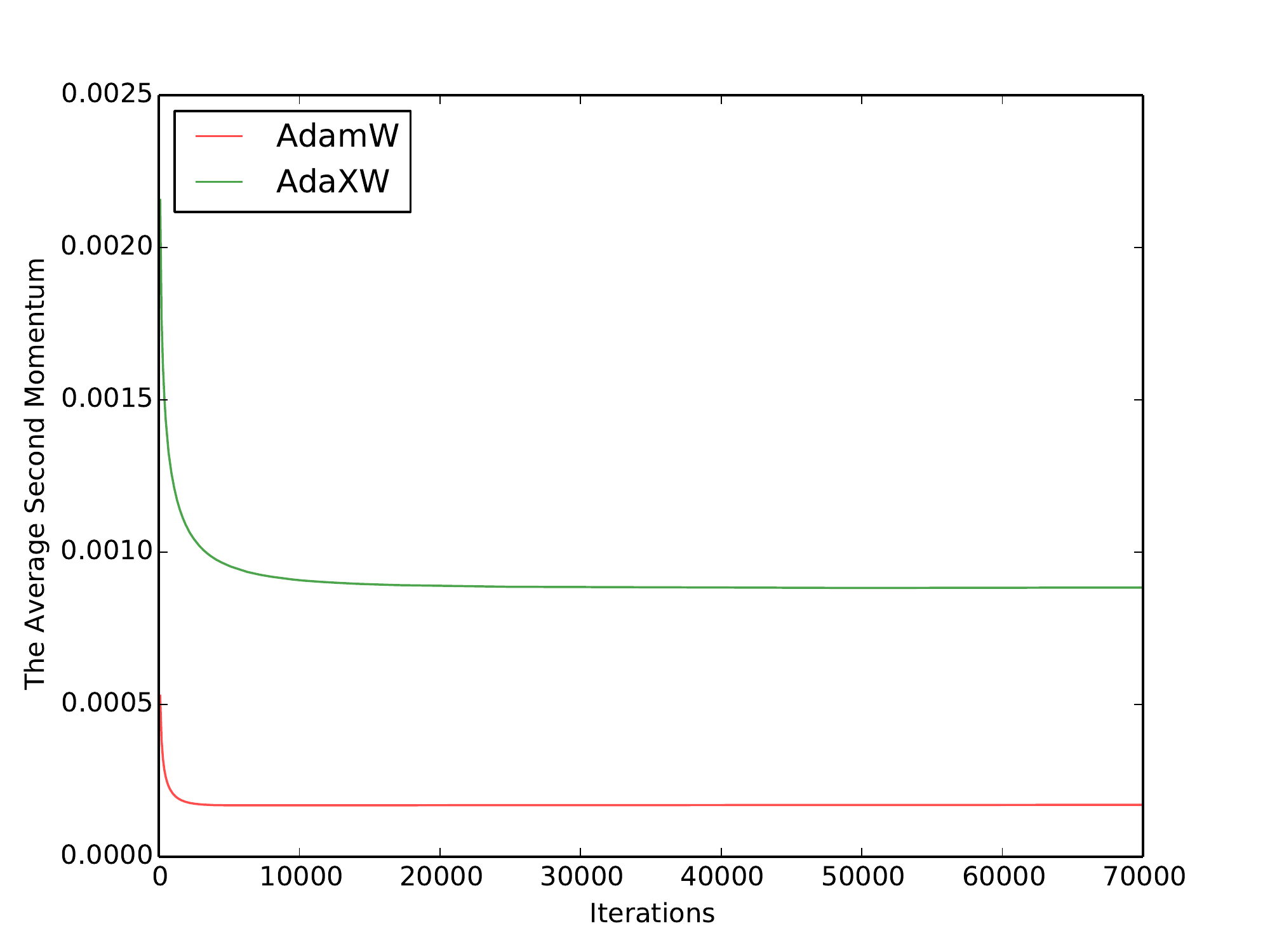}
  \label{fig: cifar_average_second}
}
\subfigure[\footnotesize Training Accuracy for Different Values of $\epsilon$]{
  \centering
  \includegraphics[width=0.23\linewidth]{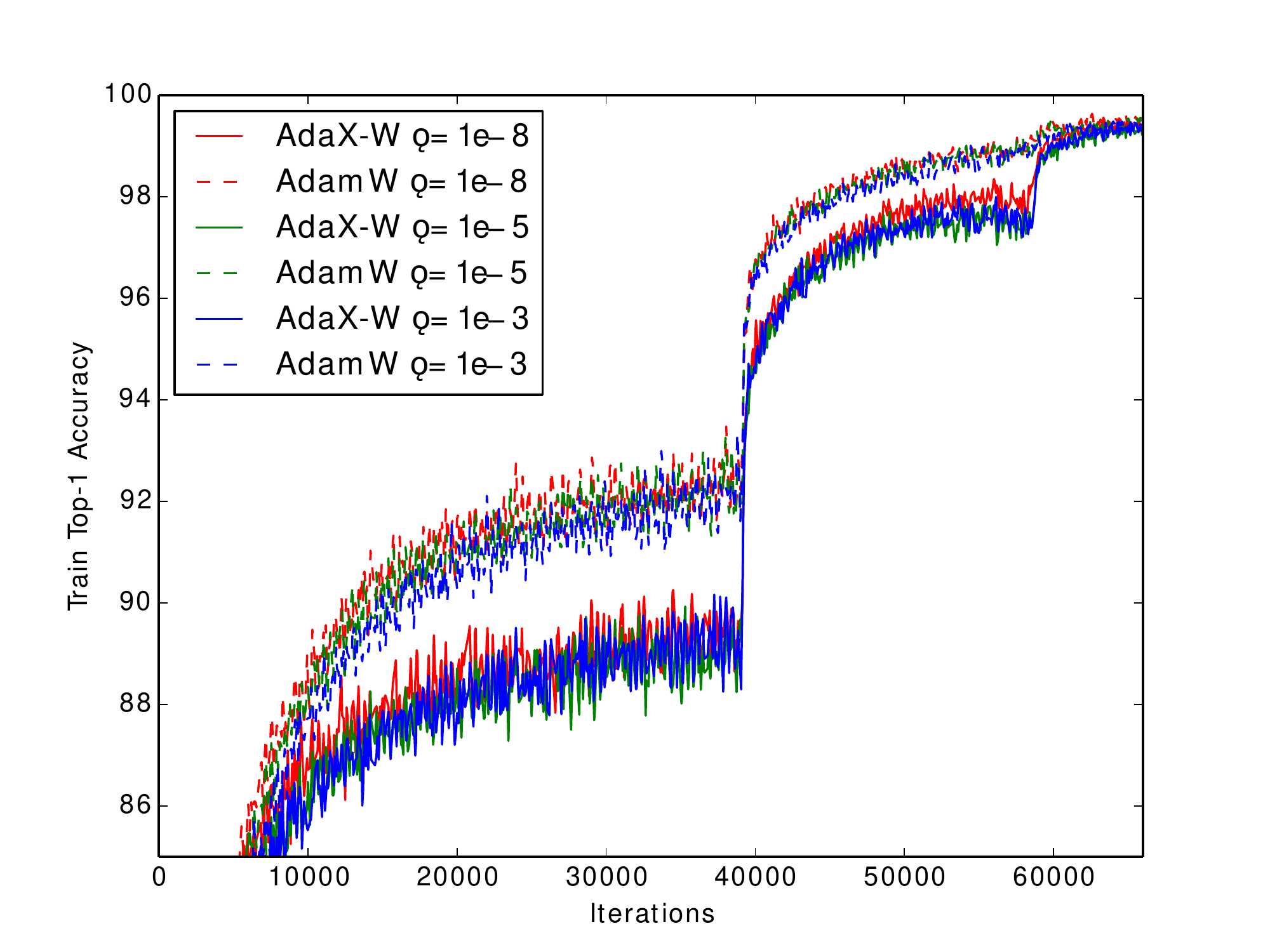}
  \label{fig: epsilon_train}
}
\subfigure[\footnotesize Testing Accuracy for Different Values of $\epsilon$]{
  \centering
  \includegraphics[width=0.23\linewidth]{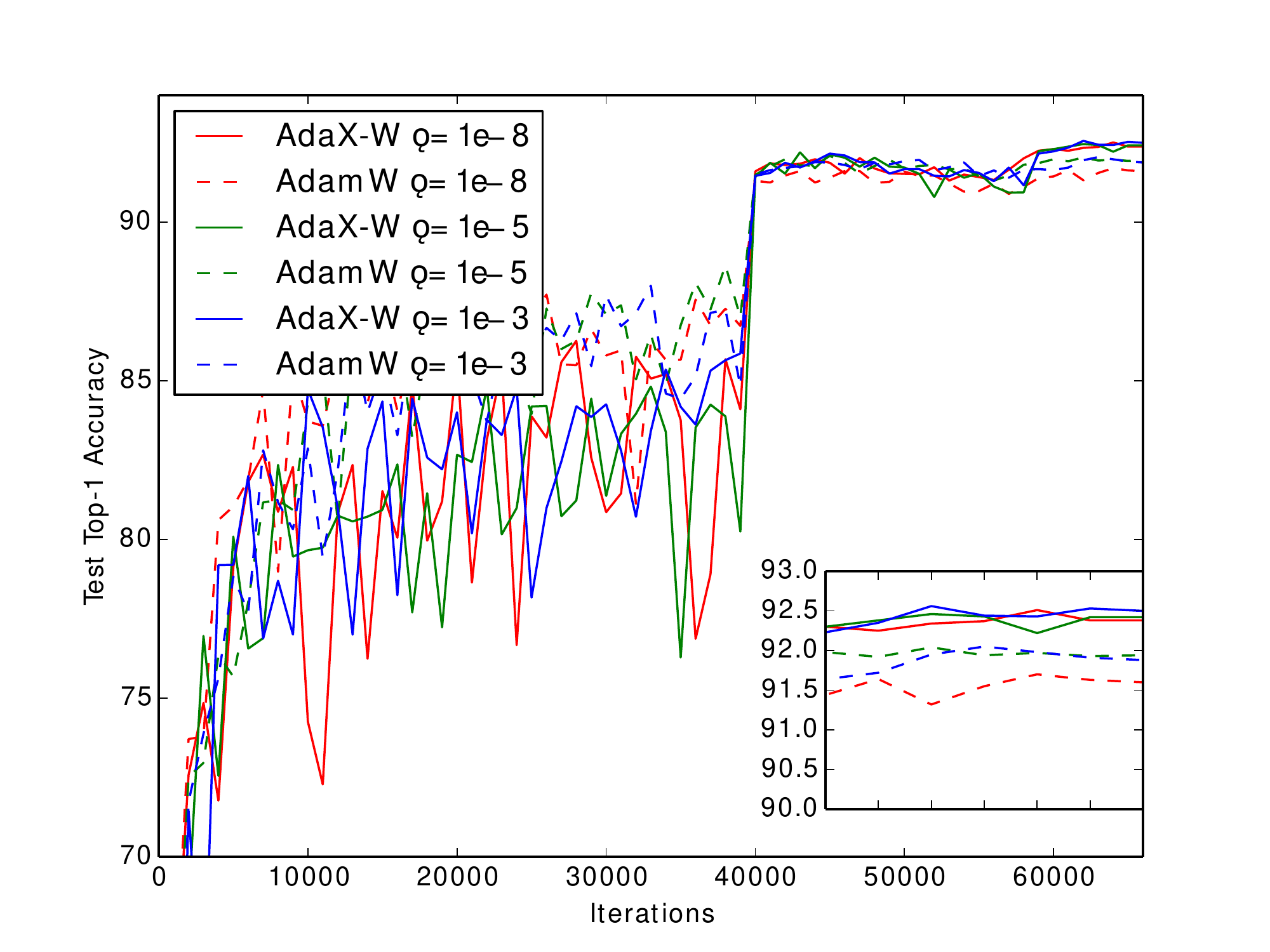}
  \label{fig: epsilon_test}
}
\caption{Second momentum stability comparisons. (a). Training updates $\alpha_t\frac{m_t}{\sqrt{\hat{v}_t}}$ for different algorithms on Problem (\ref{non_convergence_problem_2}). (b). Trace of the average second moment of AdamW and AdaX-W $\frac{\|\sqrt{\hat{v}_t}\|_1}{N}$ over iterations on CIFAR-10. (c),(d) Training and testing results of AdamW and AdaX-W with different $\epsilon$'s}
\label{fig: second_momentum_stability}
\end{figure*}

\textbf{Synthesized Example}. We first evaluated the performance of different algorithms in our synthetic problem (\ref{non_convergence_problem_2}). The problem parameters were set to be $C = 10^{-3}, \lambda = 0.9999,  x_0=1$. To ensure fair comparisons, default hyperparameters were chosen for all the algorithms, specifically $\alpha_0=0.1, \gamma=0.9$ for SGDM, $\alpha_0= 10^{-3}, \beta_1=0.9, \beta_2 = 0.999$ for Adam, and $\alpha_0= 0.005, \beta_1=0.9, \beta_2=10^{-4}$ for AdaX. As shown in Figure \ref{fig: toy_update}, SGDM and AdaX quickly converged under the strong gradient decrease information. They could potentially reach the global minimum since the change in $x_t$ remained the same regardless of initialization. However, the update steps of Adam decreased with a much slower rate, which resulted in substantial changes in $x$ and ultimately lead the algorithm to the local minimum.

\textbf{Average Second Moment}. To further prove the correctness of our theoretical findings in the synthetic problem, we tracked the average of the bias-corrected second moments in our experiments on CIFAR-10 shown in Figure \ref{fig: cifar_average_second}. It was noticeable that Adam's second moment quickly decreased to a very small number (around $2 \times 10^{-4}$) in less than 1000 iterations, showing that some of its second moments collapsed to around 0. This phenomenon revealed that the decreasing gradients in our synthetic problem (\ref{non_convergence_problem_2}) might exist in real training process. As we proved in section \ref{sec: problem}, the fast decaying second moment induced the fast convergence of Adam, but it would also possibly lead the algorithm to local minimums. In the meantime, AdaX's second moment decreased more slowly to a much larger number, which was more stable than Adam.

\textbf{Effect of \boldsymbol{$\epsilon$}}. We also performed experiments to examine AdaX's robustness with respect to different values of $\epsilon$ as discussed by \citet{zaheer2019adaptive}. In practice, people would add $\sqrt{v_t}$ by a small constant $\epsilon$ before dividing $m_t$ by $\sqrt{v_t}$ to avoid zeros in the denominators as in {Algorithm} \ref{Algorithm_detail}. However, \citet{zaheer2019adaptive} found that different values of $\epsilon$ yielded different results when using Adam. In our experiments, we found that larger values of $\epsilon \in \{1e-8, 1e-5, 1e-3\}$  improved AdamW's performance by around 0.35 percent accuracy on CIFAR-10, since it helped to stabilize very small second moment. However, AdaX-W's performance was not affected by different choices of $\epsilon$ as shown in Figure \ref{fig: epsilon_test} because its second moment was large and stable. This again proved our claim that a long-term memory design was more stable than the design of Adam.

\begin{figure}[ht] 
\centering
\subfigure[\footnotesize Padam and Padax's Training Accuracy over Iterations]{
  \centering
  \includegraphics[width=0.45\linewidth]{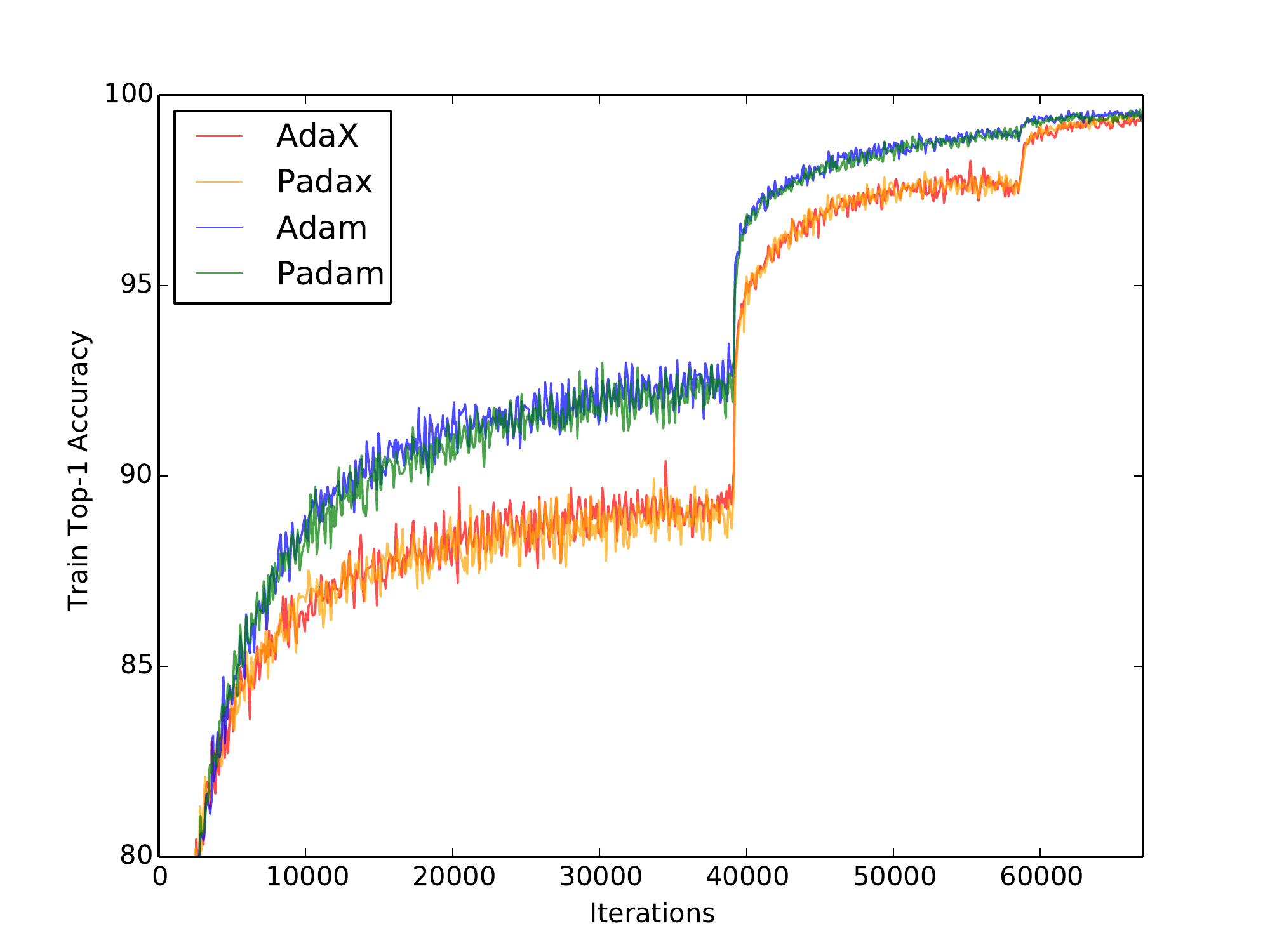}
  \label{fig: padax_train}
}
\subfigure[\footnotesize Padam and Padax's Testing Accuracy over Iterations]{
  \centering
  \includegraphics[width=0.45\linewidth]{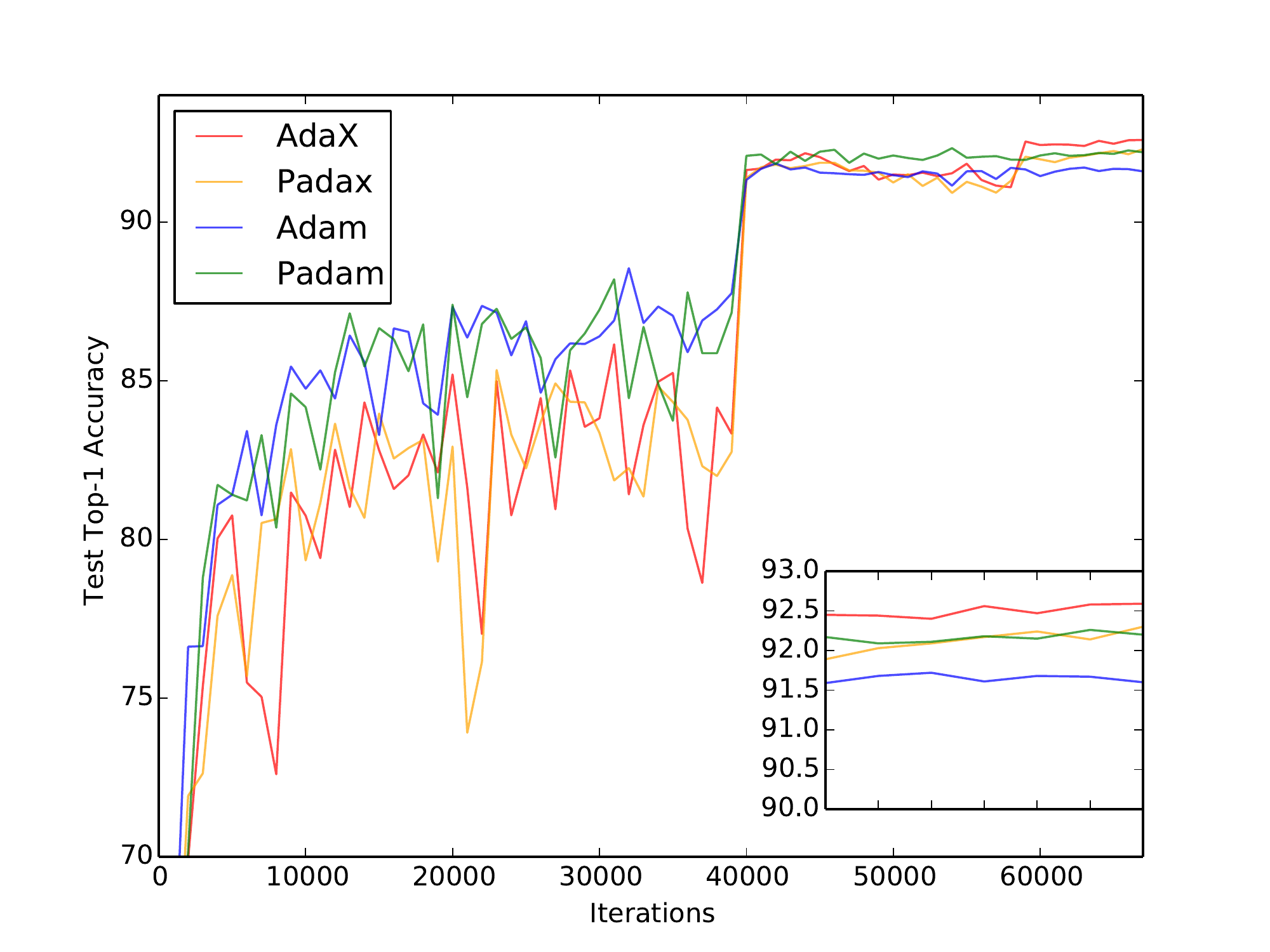}
  \label{fig: padax_test}
}
\caption{Performance of Padax and Padam on CIFAR-10}
\label{fig: padax}
\end{figure}

\subsection{Padam and Padax}

We also examined the effectiveness of changing the square root operation to $p^{th}$ power in our AdaX algorithm. We trained the Padam algorithm \citep{chen2018closing} and the corresponding Padax(Partially AdaX) algorithm on CIFAR-10 using the same settings as in subsection \ref{subsec: performance_comparison}. The best hyper-parameter $p=1/8$ as in the original paper \citep{chen2018closing} was used in our experiments. We found that changing ${V_t}^{1/2}$ to $V_t^{1/8}$ did improve Adam's performance as mentioned by  \citet{chen2018closing}. However, such modification resulted in slower training and worse testing accuracy for our method. Besides, Padam was not able to catch up with AdaX-W. We thought the reason Padam could improve over Adam was that the $(1/8)^{th}$ power could enlarge small $V_t$'s, making the algorithm less unstable and the convergence slower. However, AdaX did not have the instability problem and the modification only generated biased second moment, leading to the worse performance.

The experiments shown above verify the effectiveness of AdaX, showing that the accumulated long-term past gradient information can enhance the model performance, by getting rid of the second moment instability in vanilla Adam. It is also worth noticing that the computational cost for each step of AdaX and Adam are approximately the same, as they both memorize the first and second momentum in the past. Using the default settings, AdaX multiplies the second moment by $(1+10^{-4})$ while Adam multiplies it by $(1-10^{-3})$, but the difference is very minor. We provide the time averaged over 5 independent runs for different experiments in Table \ref{tab: time} in the Appendix. We can see that their running time are approximately the same. Therefore AdaX enables one to get higher performance than Adam in those tasks with the same training budget.

\section{Conclusion}
\label{sec: Conclusion}
In this paper, we present a novel optimization algorithm named AdaX to improve the performance of traditional adaptive methods. We first extend the non-convergence issue of Adam to a non-convex case, and show that Adam's fast convergence impairs its performance. We then propose our variant of Adam, analyze its convergence rate, and evaluate its performance on various learning tasks. Our theoretical analysis and experimental results both show that AdaX is more stable and performs better than Adam in various tasks. In the future, more experiments still need to be performed to evaluate the overall performance of AdaX and AdaX-W. Moreover, our paper is a first step into designing adaptive learning rates in ways different from simple and exponential average methods. Other new and interesting designs should also be examined. We believe that new adaptive algorithms that outperform AdaX in both convergence rate and performance still exist and remain to explore.

\bibliography{example_paper}
\bibliographystyle{icml2020}

\onecolumn
\begin{section}{Appendix}
\subsection{Proofs of Theorem \ref{Thm: Non_convergence}, Theorem \ref{Thm: Non_convergence_AMS}}
\label{App: Proofs_of_nonconvergence_problem}
We consider a one dimensional non-convex case where $\{f_t\}$ are a sequence of linear functions that have decreasing gradients in the long term. We want to show that because Adam trusts its current gradient as the second moment, its step sizes are too large and the algorithm would converge to a suboptimal solution. Let constant $C$ be the initial gradient, define cost function $f_t$ as follows:

\begin{equation}
\begin{aligned}
&f_t(x) = |C \lambda^{t-1}x|  &\forall t = 1. \\
&f_t(x) =
    \begin{cases}
     C \lambda^{t-1}x, ~~~\text{ for } x \geq 0,\\
     0, ~~~\text{ for } x < 0 ,
   \end{cases}
   &\forall t > 1.
\end{aligned}
\end{equation}

where $\lambda$ is the decreasing factor of gradient. Consider $\mathcal{F} = [-2, {C}/(1 - \lambda)]$, then it's obvious that the minimum regret is obtained at $x = 0$. Let the initial step size be $\alpha_1 = \alpha$, we then consider the performances of different algorithms in this setting. 

\textbf{(SGD)}. We first show that without the momentum, vanilla SGD is able to converge to the optimum solution and avoid going to $x<0$. Take derivative with respect to $x$, we obtain that

\begin{equation}
    \begin{aligned}
      &\nabla f_{t}(x) = C \lambda^{t-1}, \text{ for } x\geq 0 \\
      & \sum_{t=1}^{\infty} \nabla f_{t}(x)  = \sum_{t=1}^{\infty} C \lambda^{t-1} = \frac{C}{1-\lambda}
   \end{aligned}
\end{equation}

Therefore, even if we set $\alpha_t = \alpha, \forall t \geq 1$, as long as the initial point $x_0 \geq \frac{\alpha C}{1-\lambda}$, SGD is able to converge to 0. If $\alpha_t = \alpha/\sqrt{t}$, then the condition can be even less strict: $x_0 \geq \sum_{t=1}^{\infty} \frac{\alpha C \lambda^{t-1}}{\sqrt{t}}$. SGD is able to converge to the optimum if the equal signs are true.

\textbf{(Adam)}. We consider the Adam algorithm with the following parameter setting:

\begin{equation}
    \begin{aligned}
      \beta_1 = 0, 0 < \sqrt{\beta_2} < \lambda < 1, \text{ and } \alpha_t = \frac{\alpha}{\sqrt{t}}
   \end{aligned}
\end{equation}

Note that this parameter setting of Adam is the same as RMSProp, but we can further show that even if $\beta_1 \neq 0$, we still obtain similar results. Consider how $v_t$ changes in time, before it reaches the negative region, the gradients are positive and

\begin{equation}
    \begin{aligned}
      v_t &= \beta_2 v_{t-1} + (1-\beta_2)(C\lambda^{t-1})^2 \\
          &= \sum_{i=1}^t \beta_2^{t-i} (1-\beta_2)(C\lambda^{i-1})^2 \\
          &= \frac{(1-\beta_2)C^2}{\lambda^2}\beta_2^t \sum_{i=1}^t (\frac{\lambda^2}{\beta_2})^i \\
          &= \frac{(1-\beta_2)C^2 ({{\lambda^{2t}- \beta_2^t}})}{ {\lambda^2} - \beta_2}
   \end{aligned}
\end{equation}

Note that $\lambda  > \sqrt{\beta_2}$, therefore the update rule is:

\begin{equation}
    \begin{aligned}
     x_{t+1} &= x_t - \alpha_t\frac{g_t}{\sqrt{v_t}} 
     = x_t - \alpha_t\frac{\sqrt{ \lambda^2 -\beta_2}\lambda^{t-1}}{\sqrt{(1-\beta_2)( \lambda^{2t}- \beta_2^t )}} = x_t -\alpha_t\frac{\sqrt{ \lambda^2 -\beta_2}\lambda^{-1}}{\sqrt{(1-\beta_2)( 1 - (\frac{\beta_2}{\lambda^2})^t )}}\\
     &\leq x_t - \frac{\alpha}{\sqrt{t}}\sqrt{\frac{\lambda^2 -\beta_2}{\lambda^2(1-\beta_2)}}
   \end{aligned}
\end{equation}

Note that the series $\sum_{t=1}^{\infty} \frac{1}{\sqrt{t}}$ diverges, hence Adam would always reach the negative region. Same argument applies as long as $\alpha_t \geq \alpha/t$. We would emphasize here that the bias correction term in Adam does not change the final result as $ 1 - \beta_2^t \geq 1-\beta_2$ and therefore the update steps are still bounded. We could further show that when $\beta_1 \neq 0, \beta_1 < \sqrt{\beta_2}$,  Adam will still go to the negative region. Since $m_t = \beta_1 m_{t-1} + (1-\beta_1)g_t = \sum_{i=1}^t (1-\beta_1) \beta_1^{t-i}g_i$, therefore

\begin{equation}
    \begin{aligned}
     \dfrac{m_t}{\sqrt{v_t}} &= \frac{(1-\beta_1)(\lambda^t-\beta_1^t)}{\lambda-\beta_1} \cdot \frac{\sqrt{ \lambda^2 -\beta_2}}{\sqrt{(1-\beta_2)( \lambda^{2t}- \beta_2^t )}}\\
     &= \frac{(1-\beta_1)\sqrt{ \lambda^2 -\beta_2}}{(\lambda-\beta_1)\sqrt{(1-\beta_2)}} \cdot \frac{\lambda^t-\beta_1^t}{\sqrt{\lambda^{2t}- \beta_2^t}} \\
     &= \frac{(1-\beta_1)\sqrt{ \lambda^2 -\beta_2}}{(\lambda-\beta_1)\sqrt{1-\beta_2}} \cdot \frac{1 -(\beta_1/\lambda)^t}{\sqrt{1- (\beta_2/\lambda^2)^t}} \\
     &\geq \frac{(1-\beta_1)\sqrt{ \lambda^2 -\beta_2}}{(\lambda-\beta_1)\sqrt{1-\beta_2}} \cdot ({1 -\frac{\beta_1}{\lambda}})
   \end{aligned}
\end{equation}

Since the update steps are lower bounded, the algorithm would still go to the negative region.

\textbf{(AMSGrad)}. We now evaluate the performance of AMSGrad in our formulated problem. Note that $v_t$ in AMSGrad would take the same form as Adam, and $\hat{v_t} = \max\{v_t\}_{i=1}^t$. We suppose that the maximum is obtained at $v_1$ as an example, then

\begin{equation}
    \begin{aligned}
      &\hat{v}_t = {v}_{1} =  \frac{(1-\beta_2)C^2 ({{\lambda^{2}- \beta_2}})}{ \lambda^2 - \beta_2} =  (1-\beta_2)C^2  \\
      &\frac{g_t}{\sqrt{\hat{v}_t}} = \frac{g_t}{\sqrt{v_1}} = \frac{\lambda^{t-1}}{\sqrt{1-\beta_2}}\\
   \end{aligned}
\end{equation}

As we can see, AMSGrad partially solves the problem of Adam and restores the gradient decrease information as its $v_t$ is lower bounded. If the maximum of $v_t$ is obtained before the parameters enter the negative region, AMSGrad could possibly have a better performance in this problem as it prevents the update steps from being too large. However, one important determining factor is the time when the maximum value is obtained. If $v_t$ in fact keeps increasing before a very large number $T$, then AMSGrad would have the same performance as Adam. We explain the above intuition as follows. Let $h(t) = \lambda^{2t} - \beta_2^{t}$, then

\begin{equation}
    \begin{aligned}
      &\dfrac{dh(t)}{dt} = \ln{\lambda^2}
      \cdot\lambda^{2t} - \ln{\beta_2}\cdot\beta_2^{t} \\
   \end{aligned}
\end{equation}

If $\dfrac{dh(t)}{dt} \geq 0$, we have 

\begin{equation}
    \begin{aligned}
      \ln{\lambda^2}\cdot\lambda^{2t} - \ln{\beta_2}\cdot\beta_2^{t} \geq 0 \\
       (\frac{\lambda^{2}}{\beta_2})^t \leq \frac{\ln{\beta_2}}{\ln{\lambda^2}} \nonumber
   \end{aligned}
\end{equation}

When the equal sign is true, we have
\begin{equation}
    \begin{aligned}
      &t = \log_{\frac{\lambda^{2}}{\beta_2}}\frac{\ln{\beta_2}}{\ln{\lambda^2}} = \dfrac{\ln \frac{\ln{\beta_2}}{\ln{\lambda^2}}}{\ln \frac{\lambda^{2}}{\beta_2}} \\
      &\lim_{\beta_2 \rightarrow \lambda^2} t = \lim_{\beta_2 \rightarrow \lambda^2} \dfrac{\frac{1}{\beta_2\ln(\beta_2)}}{-\frac{1}{\beta_2}} = -\frac{1}{\ln(\lambda^2)}\\
      & \lim_{\lambda \rightarrow 1^{-}} t = \infty
   \end{aligned}
\end{equation}

The first equal sign in the first limit is due to L'Hospital's rule. Therefore, the value of $T$ where  $v_T = \text{max}\{v_t\}$  depends on the difference between $\beta_2$ and $\lambda^2$, and the value of $\lambda$. If $\beta_2$ is close to $\lambda^2$ or $\lambda$ is close to 1, then $v_t$ needs a large number of steps to obtain the maximum. In such cases, AMSGrad may not able to help Adam. Specifically, for a fixed $\beta_2$, since $\lim_{\lambda \rightarrow 1^{-}} t = \infty$. and 
\begin{equation}
    \begin{aligned}
    g_t/\sqrt{v_t} = \sqrt{\frac{\lambda^2 - \beta_2}{\lambda^2(1-\beta_2)}} = \frac{1}{\sqrt{1-\beta_2}} \sqrt{1 - \frac{\beta_2}{\lambda^2}}
   \end{aligned}
\end{equation}

we know that larger $\lambda^2$ will lead to both larger update steps and larger $T$ when the maximum is obtained, hence $\exists \lambda \in (0,1)$, such that AMSGrad cannot help Adam

\textbf{(AdaX)}. We provide the performance of AdaX in this problem for completeness. We only show for the case when $\beta_1 = 0$, but the same results hold when the first order momentum is used.

\begin{equation}
    \begin{aligned}
      \hat{v}_t &= \frac{1}{{(1+\beta_2)^t - 1}}\left[\sum_{i=1}^t \beta_2 (1+\beta_2)^{t-i}(C\lambda^{i-1})^2 \right]\\
          &= \frac{(1+\beta_2)^t \beta_2C^2}{{(1+\beta_2)^t - 1}} \lambda^{-2}\sum_{i=1}^t (\frac{\lambda^2}{1+\beta_2})^i \\
          &= \frac{(1+\beta_2)^t \beta_2C^2}{{(1+\beta_2)^t - 1}} \cdot \frac{1 - (\frac{\lambda^2}{1+\beta_2})^t }{{1+\beta_2}- {\lambda^2}}\\
          &= \frac{\beta_2 C^2}{{1+\beta_2}- {\lambda^2}} \cdot \frac{(1+\beta_2)^t - \lambda^{2t}}{{(1+\beta_2)^t - 1}}
   \end{aligned}
\end{equation}

\begin{equation}
    \begin{aligned}
      \frac{g_t}{\sqrt{\hat{v}_t}} 
      &= \sqrt{\frac{1+\beta_2 -\lambda^2}{\beta_2}} \cdot \frac{\lambda^{t-1}\sqrt{(1+ \beta_2)^t -1 }}{\sqrt{(1+\beta_2)^t - \lambda^{2t}}} \leq \sqrt{\frac{1+\beta_2 -\lambda^2}{\beta_2}} \lambda^{t-1}\\
   \end{aligned}
\end{equation}

As we can see,AdaX successfully restores the gradient decrease information and controls the decrease speed by an almost fixed parameter, and is therefore expected to perform better than AMSGrad since its step sizes are not affected by extreme gradients. With a suitable initial step size and starting point, AdaX is able to converge to the optimal solution 0.

\subsection{Proof of Lemma \ref{Lem: PSD_of_V}}
\textit{Proof.}
\begin{equation}
    \begin{aligned}
   \frac{V_t}{\alpha_t^2} &= \frac{t}{\alpha^2} \frac{\sum_{i=1}^t (1+\beta_2)^{t-i} \beta_2 g_i^2}{(1+ \beta_2)^t - 1} \\
        &\succeq \frac{t - 1}{\alpha^2} \frac{\sum_{i=1}^t (1+\beta_2)^{t-i} \beta_2 g_i^2}{(1+ \beta_2)^t - (1 + \beta_2)} \\
        &\succeq \frac{t - 1}{\alpha^2} \frac{\sum_{i=1}^t (1+\beta_2)^{t-i} \beta_2 g_i^2 - \beta_2 g_t^2}{(1+ \beta_2)^t - (1 + \beta_2)} \\
        &=  \frac{t - 1}{\alpha^2} \frac{\sum_{i=1}^{t-1} (1+\beta_2)^{t-1-i} \beta_2 g_i^2}{(1+ \beta_2)^{t-1} - 1}  = \frac{V_{t-1}}{\alpha_{t-1}^2} \\
   \end{aligned}
\end{equation}

where in the first inequality we utilize the fact that $(1+ \beta_2)^t \geq 1 + t\beta_2$ and hence $\frac{t}{(1+ \beta_2)^t - 1} \geq \frac{t-1}{(1+ \beta_2)^t - (1+\beta_2)}$.  Intuitively, it is easier to see this inequality if we simply let $\beta_2$ to be a small number such as 1e-4 in our implementation, then the denominator doesn't change much while the numerator decreases. 

\subsection{Auxillary Lemmas for Convergence Analysis}
\begin{lemma} Assume that $\beta_{21} = \beta_2, \beta_{2t} = \beta_2/{t}$, with $\beta_2 \in (0,1)$ and $\hat{v}_t = [(1+\beta_{2t})v_{t-1} + \beta_{2t}g_{t}^2]/[(1+\beta_{2t})^t - 1], V_t = \text{diag}(\hat{v}_t)$, then we have $\frac{V_{t}}{\alpha_t^2} \succeq \frac{V_{t-1}}{\alpha_{t-1}^2} $
\label{Lem: PSD_of_VT}
\end{lemma}
\textit{Proof: } Similar to \textbf{Lemma \ref{Lem: PSD_of_V}} in the algorithm section, we have

\begin{equation}
    \begin{aligned}
   \frac{V_t}{\alpha_t^2} &= \frac{t}{\alpha^2} \frac{\sum_{i=1}^t \beta_{2i} \Pi_{k=1}^{t-i}(1+\beta_{2(t-k+1)}) g_i^2}{(1+\beta_{2t})^t - 1} \\
        &= \frac{t}{\alpha^2} \frac{\sum_{i=1}^t \frac{\beta_{2}}{i} \Pi_{k=1}^{t-i}(1+\frac{\beta_{2}}{t-k+1}) g_i^2}{(1+\frac{\beta_{2}}{t})^t - 1}  \\
        &\succeq  \frac{t}{\alpha^2}\frac{\sum_{i=1}^{t-1} \frac{\beta_{2}}{i} \Pi_{k=1}^{t-i}(1+\frac{\beta_{2}}{t-k+1}) g_i^2}{(1+\frac{\beta_{2}}{t})^t - 1} \\
        & =  \frac{t}{\alpha^2}\frac{\sum_{i=1}^{t-1} \frac{\beta_{2}}{i} \Pi_{k=1}^{t-1-i}(1+\frac{\beta_{2}}{t-k+1}) g_i^2}{(1+\frac{\beta_{2}}{t})^{t-1} - (1+\frac{\beta_{2}}{t})^{-1}} \\
        & \succeq \frac{t-1}{\alpha^2}\frac{\sum_{i=1}^{t-1} \frac{\beta_{2}}{i} \Pi_{k=1}^{t-1-i}(1+\frac{\beta_{2}}{t-k+1}) g_i^2}{(1+\frac{\beta_{2}}{t-1})^{t-1} - 1}
        = \frac{V_{t-1}}{\alpha_{t-1}^2} \\
   \end{aligned}
\end{equation}

The first inequality comes from deleting the last term $\frac{\beta_2}{t}g_t^2$ and second one comes from the following fact: 

\begin{equation}
    \begin{aligned}
    (1+\frac{\beta_{2}}{t})^{t-1} - (1+\frac{\beta_{2}}{t})^{-1} 
    &= (1+\frac{\beta_{2}}{t})^{t-1} - \frac{t}{t+\beta_2} \\
    &\leq 1 + \frac{t-1}{t}\beta_2 + \binom{t-1}{2}(\frac{\beta_2}{t})^2 + \dots +  \binom{t-1}{t-1}(\frac{\beta_2}{t})^{t-1} - 1 + \frac{\beta_2}{t+\beta_2} \\
    & = 1 + \beta_2 + \binom{t-1}{2}(\frac{\beta_2}{t})^2 + \dots +  \binom{t-1}{t-1}(\frac{\beta_2}{t})^{t-1} - 1 + (\frac{\beta_2}{t+\beta_2} - \frac{\beta_2}{t})\\
    & \leq 1 + \beta_2 + \binom{t-1}{2}(\frac{\beta_2}{t-1})^2 + \dots +  \binom{t-1}{t-1}(\frac{\beta_2}{t-1})^{t-1} - 1 \\
    & = (1+\frac{\beta_{2}}{t-1})^{t-1} - 1
   \end{aligned}
\end{equation}

Therefore the positive semi-definiteness is satisfied.

%
%
\begin{lemma} For the parameter settings and conditions assumed in \textbf{Theorem \ref{Thm: Regret_Bound}}, we have 
\begin{equation}
    \begin{aligned}
    \sum_{t=1}^T\frac{\beta_{1t}\alpha_t}{2(1-\beta_{1t})}\|V_t^{-1/4}m_{t}\|^2 \leq \frac{\alpha C \sqrt{T}}{(1-\beta_1)^3\sqrt{\beta_2 }} \sum_{i=1}^d \|g_{1:T, i}\|_2
    \end{aligned}
\end{equation}
\label{Lem: Bound_1}
where $C$ is a constant.
\end{lemma}
\textit{Proof: } We first analyze with the following process directly from the update rules, note that 

\begin{equation}
    \begin{aligned}
    m_{T, i} &= \sum_{j=1}^T(1-\beta_{1j})\Pi_{k=1}^{T-j}\beta_{1(T-k+1)}g_{j,i} \\
    \hat{v}_{T,i} &= \frac{1}{(1+ \beta_{2T})^T-1}\sum_{j=1}^T \beta_{2j} \Pi_{k=1}^{T-j}(1+\beta_{2(T-k+1)}) g_{j,i}^2
    \end{aligned}
\end{equation}

\begin{equation}
    \begin{aligned}
& \sum_{t=1}^T\alpha_t\|V_t^{-1/4}m_{t}\|^2 \\
&= \sum_{t=1}^{T-1}\alpha_t  \|V_t^{-1/4}m_{t}\|^2  + \alpha_T \sum_{i=1}^d \frac{m^2_{T,i}}{\sqrt{\hat{v}_{T,i}}} \\
&= \sum_{t=1}^{T-1}\alpha_t  \|V_t^{-1/4}m_{t}\|^2  + \alpha \sqrt{(1+  \beta_{2T})^T-1} \sum_{i=1}^d \frac{  (\sum_{j=1}^T(1-\beta_{1j})\Pi_{k=1}^{T-j}\beta_{1(T-k+1)}g_{j,i})^2}{\sqrt{T\sum_{j=1}^T \beta_{2j} \Pi_{k=1}^{T-j}(1+\beta_{2(T-k+1)}) g_j^2}} \\
&\leq \sum_{t=1}^{T-1}\alpha_t  \|V_t^{-1/4}m_{t}\|^2  +  \alpha \sqrt{(1+  \beta_{2T})^T-1} \\
& \qquad \qquad \qquad \qquad \sum_{i=1}^d \frac{  (\sum_{j=1}^T \Pi_{k=1}^{T-j}\beta_{1(T-k+1)})(\sum_{j=1}^T (1-\beta_{1j})^2 \Pi_{k=1}^{T-j}\beta_{1(T-k+1)}g_{j,i}^2)}{\sqrt{T\sum_{j=1}^T \beta_{2j} \Pi_{k=1}^{T-j}(1+\beta_{2(T-k+1)}) g_{j,i}^2}} \\
&\leq \sum_{t=1}^{T-1}\alpha_t  \|V_t^{-1/4}m_{t}\|^2  +  \alpha \sqrt{(1+  \beta_{2T})^T-1} \sum_{i=1}^d \frac{  (\sum_{j=1}^T \beta_{1}^{T-j})(\sum_{j=1}^T \beta_{1}^{T-j}g_{j,i}^2)}{\sqrt{T{\sum_{j=1}^T \beta_{2j} \Pi_{k=1}^{T-j}(1+\beta_{2(T-k+1)}) g_j^2}}}\\
&\leq \sum_{t=1}^{T-1}\alpha_t  \|V_t^{-1/4}m_{t}\|^2  + \frac{\alpha \sqrt{(1+  \beta_{2T})^T-1} }{(1-\beta_1)\sqrt{\beta_2 T}} \sum_{i=1}^d \frac{\sum_{j=1}^T \beta_{1}^{T-j}g_{j,i}^2}{\sqrt{{\sum_{j=1}^T\frac{1}{j}\Pi_{k=1}^{T-j}(1+\beta_{2(T-k+1)}) g_{j,i}^2}}}\\
&\leq  \sum_{t=1}^{T-1}\alpha_t  \|V_t^{-1/4}m_{t}\|^2  + \frac{\alpha \sqrt{(1+  \beta_{2T})^T-1} }{(1-\beta_1)\sqrt{\beta_2 T}} \sum_{i=1}^d \sum_{j=1}^T\frac{ \sqrt{j} \beta_{1}^{T-j}g_{j,i}^2}{\sqrt{{\Pi_{k=1}^{T-j}(1+\beta_{2(T-k+1)})  g_{j,i}^2}}}\\
&\leq \sum_{t=1}^{T-1}\alpha_t  \|V_t^{-1/4}m_{t}\|^2  + \frac{\alpha C }{(1-\beta_1)\sqrt{\beta_2 T}} \sum_{i=1}^d \sum_{j=1}^T { \beta_{1}^{T-j}|g_{j,i}|}\sqrt{j}\\
    \end{aligned}
\end{equation}

where the first inequality is due to an application of Cauchy-Schwarz inequality. The second inequality is due to the fact that $\beta_{1t} \leq \beta_1, \forall t$. The third inequality follows from $\sum_{j=1}^T \beta_1^{T-j} \leq 1/(1-\beta_1)$ and the fact that $1-\beta_{1j} \leq 1$. The fourth one comes from only keeping one of the positive terms in the denominator. The final one is from the fact that $\sqrt{(1+  \beta_{2T})^T-1} \leq C$ for a constant $C = \sqrt{e^{\beta_2} - 1}$ and $\beta_{2j}\geq 0$. By using induction on all the terms in the equation, we are able to further bound it.

\begin{equation}
    \begin{aligned}
 \sum_{t=1}^T\alpha_t\|V_t^{-1/4}m_{t}\|^2 
&\leq \sum_{t=1}^{T}\frac{\alpha C }{(1-\beta_1)\sqrt{\beta_2 t}} \sum_{i=1}^d \sum_{j=1}^t  \beta_{1}^{t-j}|g_{j,i}|\sqrt{j}
= \sum_{t=1}^{T}\frac{\alpha C }{(1-\beta_1)\sqrt{\beta_2}} \sum_{i=1}^d \sum_{j=1}^t \sqrt{\frac{j}{t}}{ \beta_{1}^{t-j}|g_{j,i}|} \\
&= \frac{\alpha C }{(1-\beta_1)\sqrt{\beta_2 }} \sum_{i=1}^d \sum_{t=1}^{T} |g_{t,i}| \sum_{j=t}^T  \sqrt{\frac{t}{j}}{ \beta_{1}^{j-t}} 
\leq \frac{\alpha C }{(1-\beta_1)\sqrt{\beta_2 }} \sum_{i=1}^d \sum_{t=1}^{T} |g_{t,i}| \sum_{j=t}^T { \beta_{1}^{j-t}} \\
& \leq \frac{\alpha C }{(1-\beta_1)^2\sqrt{\beta_2 }} \sum_{i=1}^d \sum_{t=1}^{T} |g_{t,i}| \leq \frac{\alpha C }{(1-\beta_1)^2\sqrt{\beta_2 }} \sum_{i=1}^d \sqrt{\sum_{t=1}^{T} |g_{t,i}|^2} \sqrt{\sum_{t=1}^T 1}\\
&= \frac{\alpha C \sqrt{T}}{(1-\beta_1)^2\sqrt{\beta_2 }} \sum_{i=1}^d \sqrt{\sum_{t=1}^{T} |g_{t,i}|^2}
    \end{aligned}
\end{equation}

The second equality follows from a re-arrange of sum order. The third inequality comes from the fact that $\sum_{j=t}^T \beta_1^{j-t} \leq 1/(1-\beta_1) \leq 1/(1-\beta_1)$. The final inequality is again from the Cauchy-Schwarz inequality. By denoting $g_{1:t, i}$ to be the vector of the past gradients from time $1$ to $t$ in the $i$-th dimension, i.e. $g_{1:t, i} = [g_{1,i}, g_{2,i}, ..., g_{t,i}]$, and by the fact that $\beta_1/2 <1$, we complete the proof of the lemma.

%
%
\begin{lemma} For the parameter settings and conditions assumed in \textbf{Theorem \ref{Thm: Regret_Bound}}, we have 
\begin{equation}
    \begin{aligned}
    \sum_{t=1}^T \frac{1}{\alpha_t}\left[\| V_t^{{1/4}}(x_{t} - x^*) \|^2 - \| V_t^{{1/4}}(x_{t+1} - x^*) \|^2 \right] \leq \frac{D_{\infty}^2}{2\alpha_T} \sum_{i=1}^d \hat{v}_{T, i}^{1/2}
    \end{aligned}
\end{equation}
\label{Lem: Bound_2}
\end{lemma}
\textit{Proof: } Using the definition of $L2$ norm, by \textbf{Lemma \ref{Lem: PSD_of_VT}}, since $\frac{\hat{v}_{t,i}^{1/2}}{\alpha_t} \geq  \frac{\hat{v}_{t-1,i}^{1/2}}{\alpha_{t-1}} $

\begin{equation}
    \begin{aligned}
        &\sum_{t=1}^T \frac{1}{\alpha_t}\left[\| V_t^{{1/4}}(x_{t} - x^*) \|^2 - \| V_t^{{1/4}}(x_{t+1} - x^*) \|^2 \right] \\
        &\leq  \frac{1}{\alpha_1} \| V_1^{{1/4}}(x_{1} - x^*) \|^2 + 
        \sum_{t=2}^T \left[\frac{\| V_t^{{1/4}}(x_{t} - x^*) \|^2}{\alpha_t} - \frac{\| V_{t-1}^{{1/4}}(x_{t} - x^*) \|^2}{\alpha_{t-1}} \right]  \\
        &=  \frac{1}{\alpha_1} \sum_{i=1}^d \hat{v}_{1,i}^{1/2}(x_{1,i} - x_{i}^*)^2 + 
        \sum_{t=2}^T \sum_{i=1}^d \left[ \frac{\hat{v}_{t,i}^{1/2}}{\alpha_t}(x_{t,i} - x_{i}^*)^2 - \frac{\hat{v}_{t-1,i}^{1/2}}{\alpha_{t-1}}(x_{t,i} - x_{i}^*)^2 \right] \\
        & = \frac{1}{\alpha_1} \sum_{i=1}^d \hat{v}_{1,i}^{1/2}(x_{1,i} - x_{i}^*)^2 + 
        \sum_{t=2}^T \sum_{i=1}^d \left[ \frac{\hat{v}_{t,i}^{1/2}}{\alpha_t} - \frac{\hat{v}_{t-1,i}^{1/2}}{\alpha_{t-1}} \right](x_{t,i} - x_{i}^*)^2 \\
        &\leq \frac{D_{\infty}^2}{2\alpha_T} \sum_{i=1}^d \hat{v}_{T, i}^{1/2} 
    \end{aligned}
\end{equation}

where the first inequality is from separating the first term and getting rid of the last negative term in the summation. The last inequality is from a telescopic summation and the diameter bound that $ \|x-x^*\| \leq D_{\infty}$

%
%
\subsection{Proof of Regret Bound}
\subsubsection{Proof of Theorem \ref{Thm: Regret_Bound}}
\textit{Proof}. Following the proof given by \citet{Reddi2018On}, we provide the proof of regret bound in \textbf{Theorem \ref{Thm: Regret_Bound}}. Beginning with the definition of the projection operation $\Pi_{\mathcal{F}, \sqrt{V_t}}$, we have the observation

\begin{equation}
    \begin{aligned}
        x_{t+1} = \Pi_{\mathcal{F}, \sqrt{V_t}}(x_{t} - \alpha_t V_t^{-{1/2}}m_t) = \min_{x \in \mathcal{F}} \| V_t^{{1/4}}(x-(x_{t} - \alpha_t V_t^{-{1/2}}m_t)) \|
    \end{aligned}
\end{equation}

Using Lemma 4 in \citet{Reddi2018On} proved by \citet{Mcmahan2010Adaptive}\ with a direct substitute of $z_1 = (x_{t} - \alpha_t V_t^{-{1/2}}m_t), Q = V^{1/2}$ and $z_2 = x^*$ for $x^* \in \mathcal{F}$, the following inequality holds: 

\begin{equation}
    \begin{aligned}
        \| V_t^{{1/4}}(u_1 - u_2) \|^2
        &= \| V_t^{{1/4}}(x_{t+1} - x^*) \|^2 
        \leq \| V_t^{{1/4}}(x_{t} - \alpha_t V_t^{-{1/2}}m_t - x^*) \|^2 \\
        &= \| V_t^{{1/4}}(x_{t} - x^*) \|^2 + \alpha_t^2 \|V_t^{-{1/4}}m_t \|^2 - 2\alpha_t \langle m_t, (x_t - x^*)\rangle \\
        &= \| V_t^{{1/4}}(x_{t} - x^*) \|^2 + \alpha_t^2 \|V_t^{-{1/4}}m_t \|^2 - 2\alpha_t \langle \beta_{1t}m_{t-1} + (1-\beta_{1t}) g_t, (x_t - x^*)\rangle \\
    \end{aligned}
\end{equation}

where the first equality is due to the fact that $\Pi_{\mathcal{F}, \sqrt{V_t}}(x^*) = x^*$. Rearrange the last inequality, we obtain

\begin{equation}
    \begin{aligned}
        (1-\beta_{1t})\langle  g_t, (x_t - x^*)\rangle 
        &\leq \frac{1}{2\alpha_t}\left[\| V_t^{{1/4}}(x_{t} - x^*) \|^2 - \| V_t^{{1/4}}(x_{t+1} - x^*) \|^2 \right] +  \frac{\alpha_t}{2} \|V_t^{-{1/4}}m_t \|^2 \\
        &\qquad \qquad - \beta_{1t}\langle m_{t-1}, (x_t - x^*)\rangle  \\
        & \leq \frac{1}{2\alpha_t}\left[\| V_t^{{1/4}}(x_{t} - x^*) \|^2 - \| V_t^{{1/4}}(x_{t+1} - x^*) \|^2 \right] +  \frac{\alpha_t}{2} \|V_t^{-{1/4}}m_t \|^2 \\
        &\qquad \qquad + \frac{\beta_{1t}\alpha_t}{2}\|V_t^{-1/4}m_{t-1}\|^2 +  \frac{\beta_{1t}}{2\alpha_t}\|V_t^{1/4}(x_t - x^*)\|^2  \\
    \end{aligned}
\end{equation}

The second inequality comes from applications of Cauchy-Schwarz and Young's inequality. We now make use of the approach of bounding the regret using convexify of $f_t$ as in \citet{Kingma2015Adam}. Following \textbf{Lemma \ref{Lem: Bound_1}} and \textbf{Lemma \ref{Lem: Bound_2}}, we have

\begin{equation}
    \begin{aligned}
        &\sum_{t=1}^T f_t (x_t) -f_t (x^*) 
        \leq \sum_{t=1}^T \langle  g_t, (x_t - x^*)\rangle \\
        & \leq \sum_{t=1}^T \frac{1}{2\alpha_t(1-\beta_{1t})}\left[\| V_t^{{1/4}}(x_{t} - x^*) \|^2 - \| V_t^{{1/4}}(x_{t+1} - x^*) \|^2 \right] +  \frac{\alpha_t}{2(1-\beta_{1t})} \|V_t^{-{1/4}}m_t \|^2 \\
        &\qquad \qquad + \frac{\beta_{1t}\alpha_t}{2(1-\beta_{1t})}\|V_t^{-1/4}m_{t-1}\|^2 +  \frac{\beta_{1t}}{2\alpha_t(1-\beta_{1t})}\|V_t^{1/4}(x_t - x^*)\|^2  \\
        & \leq \frac{D_{\infty^2}}{2\alpha_T(1-\beta_1)}\sum_{i=1}^d \hat{v}_{T,i}^{1/2} + \sum_{t=1}^T \frac{\beta_{1t}}{2\alpha_t(1-\beta_{1})}\|V_t^{1/4}(x_t - x^*)\|^2 + \frac{\alpha C \sqrt{T}}{(1-\beta_1)^3\sqrt{\beta_2 }} \sum_{i=1}^d \|g_{1:T, i}\|_2 \\
        & = \frac{D_{\infty^2}}{2\alpha_T(1-\beta_1)}\sum_{i=1}^d \hat{v}_{T,i}^{1/2} + \sum_{t=1}^T  \frac{1}{2\alpha_t(1-\beta_{1})}\sum_{i=1}^d \beta_{1t}(x_{t,i} -x_{i}^*)^2 \hat{v}_{t,i}^{1/2} + \frac{\alpha C \sqrt{T}}{(1-\beta_1)^3\sqrt{\beta_2 }} \sum_{i=1}^d \|g_{1:T, i}\|_2 \\
        & \leq \frac{D_{\infty^2}}{2\alpha_T(1-\beta_1)}\sum_{i=1}^d \hat{v}_{T,i}^{1/2} + \frac{D_{\infty}^2}{2(1-\beta_1)} \sum_{t=1}^T  \sum_{i=1}^d \frac{\beta_{1t}\hat{v}_{t,i}^{1/2}}{\alpha_t} + \frac{\alpha C \sqrt{T}}{(1-\beta_1)^3\sqrt{\beta_2 }} \sum_{i=1}^d \|g_{1:T, i}\|_2 \\
        & \leq \frac{D_{\infty^2}}{2\alpha_T(1-\beta_1)}\sum_{i=1}^d \hat{v}_{T,i}^{1/2} + \frac{D_{\infty}^2}{2(1-\beta_1)} \sum_{t=1}^T  \sum_{i=1}^d \frac{\beta_{1t}\hat{v}_{t,i}^{1/2}}{\alpha_t} + \frac{\alpha C \sqrt{T}}{(1-\beta_1)^3\sqrt{\beta_2 }} \sum_{i=1}^d \|g_{1:T, i}\|_2 \\
    \end{aligned}
\end{equation}

\subsubsection{Proof of Corollary \ref{Cor: Regret_Bound}}
\textit{Proof. } We first take a look at the size of ${\hat{v}_{t,i}}^{1/2}$, note that $\|\nabla f_t(\theta)\|_{\infty} \leq G_{\infty}$

\begin{equation}
    \begin{aligned}
         \hat{v}_{t,i} &= \frac{1}{(1+ \beta_{2t})^t-1}\sum_{j=1}^t \beta_{2j} \Pi_{k=1}^{t-j}(1+\beta_{2(t-k+1)}) g_{j,i}^2 \\
        &\leq \frac{G_{\infty}^2}{\beta_{2}}\sum_{j=1}^t \frac{\beta_{2}}{j} \Pi_{k=1}^{t-j}(1+\frac{\beta_{2}}{t-k+1}) \\ 
        & \leq G_{\infty}^2 \sum_{j=1}^t \frac{1}{j} \Pi_{k=1}^{t-j}(1+\frac{1}{t-k+1}) \\
        & = G_{\infty}^2 \sum_{j=1}^t \frac{t+1}{j(j+1)} = t G_{\infty}^2\\
    \end{aligned}
\end{equation}

The first inequality is due to the fact that $ (1+ \beta_{2t})^t \geq (1+ \beta_2)$ and the gradient bound. The second inequality follows from $\beta_2 < 1$. The last inequality is from the telescopic sum. Then we have the following inequality,

\begin{equation}
    \begin{aligned}
          \sum_{t=1}^T  \sum_{i=1}^d \frac{\beta_{1t}\hat{v}_{t,i}^{1/2}}{\alpha_t} 
          \leq d G_{\infty} \frac{\beta_1}{\alpha} \sum_{t=1}^{T}\lambda^{t-1}t
          \leq \frac{d G_{\infty}\beta_1}{\alpha(1-\lambda)^2}
    \end{aligned}
\end{equation}

The second inequality is due to the arithmetic geometric series sum $\sum_{t=1}^{T}\lambda^{t-1}t < \frac{1}{(1-\lambda)^2}$, the reason is as follows

\begin{equation}
    \begin{cases}
          S = \lambda^0 + 2\lambda^1 + \cdots + t\lambda^{t-1} \\
          \lambda S = \lambda^1 + 2\lambda^2 + \cdots + t\lambda^{t} \\
    \end{cases}
\end{equation}

\begin{equation}
    \begin{aligned}
    (1-\lambda)S = \lambda^0 + \lambda^1 + \cdots + \lambda^{t-1} - t\lambda^t
    \leq \lambda^0 + \lambda^1 + \cdots + \lambda^{t-1}
     \leq \frac{1}{1-\lambda}
    \end{aligned}
\end{equation}

Therefore we have the following regret bound 

\begin{equation}
    \begin{aligned}
   R_T & \leq \frac{D_{\infty^2}\sqrt{T}}{2\alpha(1-\beta_1)}\sum_{i=1}^d \hat{v}_{T,i}^{1/2} + \frac{d\beta_1D_{\infty}^2 G_{\infty}}{2\alpha(1-\beta_1)(1-\lambda)^2} + \frac{\alpha C \sqrt{T}}{(1-\beta_1)^3\sqrt{\beta_2 }} \sum_{i=1}^d \|g_{1:T, i}\|_2 \\
   \end{aligned}
\end{equation}

\subsection{Proof of Non-Convex Convergence Rate}

\subsubsection{Proof of Theorem \ref{Thm: Stationary}}
\textit{Proof.} We first directly refer to the original paper and obtain the following bound \cite{Chen2019On}.
\begin{equation}
    \begin{aligned}
    &\mathbb{E}\left[ \sum_{t=1}^T \alpha_t \langle \nabla f(x_t), \nabla f(x_t)/\sqrt{\hat{v}_t}\rangle \right]\\
    &\leq \mathbb{E} \left[ C_1\sum_{t=1}^T\| \alpha_t g_t/\sqrt{v_t}\|^2 + C_2\sum_{t=2}^{T}\| \frac{\alpha_t}{\sqrt{\hat{v}_t}} -  \frac{\alpha_{t-1}}{\sqrt{\hat{v}_{t-1}}}\|_1 + C_3\sum_{t=2}^{T-1}\| \frac{\alpha_t}{\sqrt{\hat{v}_t}} - \frac{\alpha_{t-1}}{\sqrt{\hat{v}_{t-1}}}\|^2 \right] + C_4\\
   \end{aligned}
\end{equation}

where $C_1, C_2, C_3$ are constants independent of $d$ and $T$, $C_4$ is a constant independent of $T$. For the first term, assume that $\text{min}_{j\in [d]} (\sqrt{\hat{v}_1})_j \geq c > 0$, we have

\begin{equation}
    \begin{aligned}
 \qquad \mathbb{E} \left[ C_1\sum_{t=1}^T\| \alpha_t g_t/\sqrt{v_t}\|^2 \right] &
 \leq \mathbb{E} \left[ C_1\sum_{t=1}^T\| \alpha_t g_t/c\|^2 \right] 
 = \mathbb{E} \left[ C_1\sum_{t=1}^T \frac{\alpha^2}{c^2t}\| g_t\|^2 \right]\\
&\leq  \frac{C_1G_{\infty}^2 \alpha^2}{c^2}(1+\log T)
   \end{aligned}
\end{equation}

where the first inequality follows from \textbf{Lemma \ref{Lem: PSD_of_VT}} as $\frac{\hat{v}_t}{\alpha_t^2} \geq \frac{\hat{v}_{t-1}}{{\alpha_t^2}}$. The second inequality is from the gradient bound $\|\nabla f(x_t)\| \leq G_{\infty}$. The last inequality is due to the harmonic series $\sum_{t=1}^N \frac{1}{t} \leq (1+\log T)$. For the second term with $C_2$, similarly by the positive semi-definiteness in \textbf{Lemma \ref{Lem: PSD_of_VT}}, we have

\begin{equation}
    \begin{aligned}
& \qquad \mathbb{E}\left[C_2\sum_{t=2}^{T}\| \frac{\alpha_t}{\sqrt{\hat{v}_t}} -  \frac{\alpha_{t-1}}{\sqrt{\hat{v}_{t-1}}}\|_1 \right] =  \mathbb{E}\left[C_2\sum_{j=1}^d \sum_{t=2}^{T}\left(\frac{\alpha_{t-1}}{(\sqrt{\hat{v}_{t-1}}})_j - \frac{\alpha_t}{(\sqrt{\hat{v}_t})_j}\right) \right] \\
&= \mathbb{E}\left[C_2\sum_{j=1}^d \left(\frac{\alpha_{1}}{(\sqrt{\hat{v}_{1}})_j} - \frac{\alpha_T}{(\sqrt{\hat{v}_T})_j}\right) \right] \leq \mathbb{E}\left[C_2\sum_{j=1}^d \frac{\alpha_{1}}{(\sqrt{\hat{v}_{1}})_j} \right] \leq \frac{C_2 d \alpha}{c}
   \end{aligned}
\end{equation}

The second equality is from the telescope sum and for the third term

\begin{equation}
    \begin{aligned}
& \mathbb{E}\left[C_3\sum_{t=2}^{T-1}\| \frac{\alpha_t}{\sqrt{\hat{v}_t}} -  \frac{\alpha_{t-1}}{\sqrt{\hat{v}_{t-1}}}\|^2 \right] \leq \mathbb{E}\left[C_3\sum_{t=2}^{T-1}\frac{\alpha}{c}\| \frac{\alpha_t}{\sqrt{\hat{v}_t}} -  \frac{\alpha_{t-1}}{\sqrt{\hat{v}_{t-1}}}\|_1 \right] \leq  \frac{C_3 d \alpha^2}{c^2}
   \end{aligned}
\end{equation}

where the first inequality is because $|\frac{\alpha_t}{\sqrt{v_t}} - \frac{\alpha_{t-1}}{\sqrt{v_{t-1}}}| \leq \frac{\alpha_{t-1}}{\sqrt{v_{t-1}}} \leq \frac{\alpha}{c}$ and the last one is due to the previous inequality with second term. Hence in summary, we have 

\begin{equation}
    \begin{aligned}
    \mathbb{E}\left[ \sum_{t=1}^T \alpha_t \langle \nabla f(x_t), \nabla f(x_t)/\sqrt{\hat{v}_t}\rangle \right]
    \leq \frac{C_1G_{\infty}^2 \alpha^2}{c^2}(1+\log T) + \frac{C_2 d \alpha}{c} + \frac{C_3 d \alpha^2}{c^2} + C_4
   \end{aligned}
\end{equation}

Note that $(\hat{v}_t)_j$ has the following upper bound as $\|\nabla f_t(\theta)\|_{\infty} \leq G_{\infty}$,

\begin{equation}
    \begin{aligned}
         \hat{v}_{t,i} &= \frac{1}{(1+ \beta_{2})^t-1}\sum_{j=1}^t \beta_{2} (1+\beta_2)^{t-j} g_{j,i}^2 \\
        &\leq \frac{G_{\infty}^2}{(1+ \beta_{2})^t-1}\sum_{j=1}^t \beta_{2} (1+\beta_2)^{t-j} = G_{\infty}^2
    \end{aligned}
\end{equation}

And thus we have 

\begin{equation}
    \begin{aligned}
    \mathbb{E}\left[ \sum_{t=1}^T \alpha_t \langle \nabla f(x_t), \nabla f(x_t)/\sqrt{\hat{v}_t}\rangle \right]
    &\geq \mathbb{E}\left[ \sum_{t=1}^T \frac{\alpha}{\sqrt{t} G_{\infty}} \|\nabla f(x_t)\|^2 \right] \geq  \frac{\alpha}{ G_{\infty}} \min_{t\in [T]}\mathbb{E}\left[ \|\nabla f(x_t)\|^2 \right] \sum_{t=1}^T \frac{1}{\sqrt{t}} \\
    &\geq  \frac{\alpha}{ G_{\infty}} \min_{t\in [T]}\mathbb{E}\left[ \|\nabla f(x_t)\|^2 \right] \sqrt{T}
   \end{aligned}
\end{equation}

where the last inequality is by the fact that $\sum_{t=1}^T \frac{1}{\sqrt{t}} \geq \sqrt{T}$, therefore we have

\begin{equation}
    \begin{aligned}
 \min_{t\in [T]}\mathbb{E}\left[ \|\nabla f(x_t)\|^2 \right] \leq \frac{G_{\infty}}{\alpha \sqrt{T}} (\frac{C_1G_{\infty}^2 \alpha^2}{c^2}(1+\log T) + \frac{C_2 d \alpha}{c} + \frac{C_3 d \alpha^2}{c^2} + C_4)
   \end{aligned}
\end{equation}

We would emphasize that the assumption $\|\alpha_t m_t /\sqrt{\hat{v}_t}\| \leq G $ in the theorem is automatically satisfied as $\frac{\alpha_{t}}{\sqrt{\hat{v}_{t}}} \leq \frac{\alpha_{1}}{\sqrt{\hat{v}_{1}}} = \frac{\alpha}{c}$. Hence  $\|\alpha_t m_t /\sqrt{\hat{v}_t}\| \leq  \frac{\alpha G_{\infty}}{c}$.

\subsection{Implementation Details}
\label{App: implementation}
The detailed implementations of AdaX with $L_2$ regularization and AdaX-W are as in Algorithm \ref{Algorithm_detail}. The performance of AdaX is robust with respect to the value of $\beta_2$, but we recommend smaller values such as $1e-4, 1e-5$ to reduce computational cost. Note that the main differences between AdaX and AdaXW are in line 4 and line 9, where $L_2$ regularization and decoupled weight decay are applied. The small constant $\epsilon$ in line 7 is used to avoid zeros in the denominators and we have shown the choice of $\epsilon$ does not affect the performance of AdaX in section \ref{sec: Experiments}.

\begin{algorithm}[H]
   \caption{AdaX Algorithm with \colorbox{yellow}{$L_2$  Regularization} and \colorbox{green}{Decoupled Weight Decay}}
   \label{Algorithm_detail}
\begin{algorithmic}
   \STATE \textbf{Input:} Initialize $x_0$, step size $\{\alpha_t\}_{t=1}^{T}, (\beta_{1}, \beta_{2}) = (0.9, 1e-4), \text{weight decay } \lambda $, $\epsilon = 1e-12$ 
   \STATE \textbf{Initialize} $m_0 = 0, v_0 = 0$
   \FOR{$ t= 1 $ \textbf{to} $T$}
   \STATE $g_t = \nabla f_t(x_t)$ \colorbox{yellow}{+ $\lambda x_t$}
   \STATE $m_t =  \beta_{1}m_{t-1} + (1-\beta_1) g_t$
   \STATE $v_t =  (1+\beta_{2})v_{t-1} + \beta_{2} g_t^2 $
   \STATE $d_t = \sqrt{v_t} + \epsilon$
   \STATE $\hat{d}_t = d_t / \sqrt{(1+\beta_{2})^t -1}$ and $V_t = \text{diag}(\hat{d}_t)$
   \STATE  $x_{t+1} = \Pi_{\mathcal{F}, {V_t}}(x_t - \alpha_t (m_t/{\hat{d}_t}) \colorbox{green}{+ $\alpha_t\lambda x_t$})$
   \ENDFOR
\end{algorithmic}
\end{algorithm}

\textbf{Comparison Between $L_2$ Regularization and Decoupled Weight Decay},.
We also compared the differences between $L_2$-regularization and decoupled weight decay in our AdaX algorithm as in \citet{Loshchilov2019Decoupled}. We trained the ResNet-18 model on CIFAR-10 with AdaX and Adam using $L_2$ regularization and decoupled weight decay. As shown in Figure \ref{fig: L2-WD}, Adam and AdaX with decoupled weight decay (AdamW, AdaX-W) performed much better in both the training and the testing stages, and therefore decoupled weight decay is better than $L_2$ regularization. It was worth noticing that AdaX also performed better than Adam when using $L_2$ regularization, which proved our claim that AdaX was a better choice than Adam.

\begin{figure}[H]
\centering
\subfigure[Training Top-1 Accuracy]{
  \centering
  \includegraphics[width=0.4\linewidth]{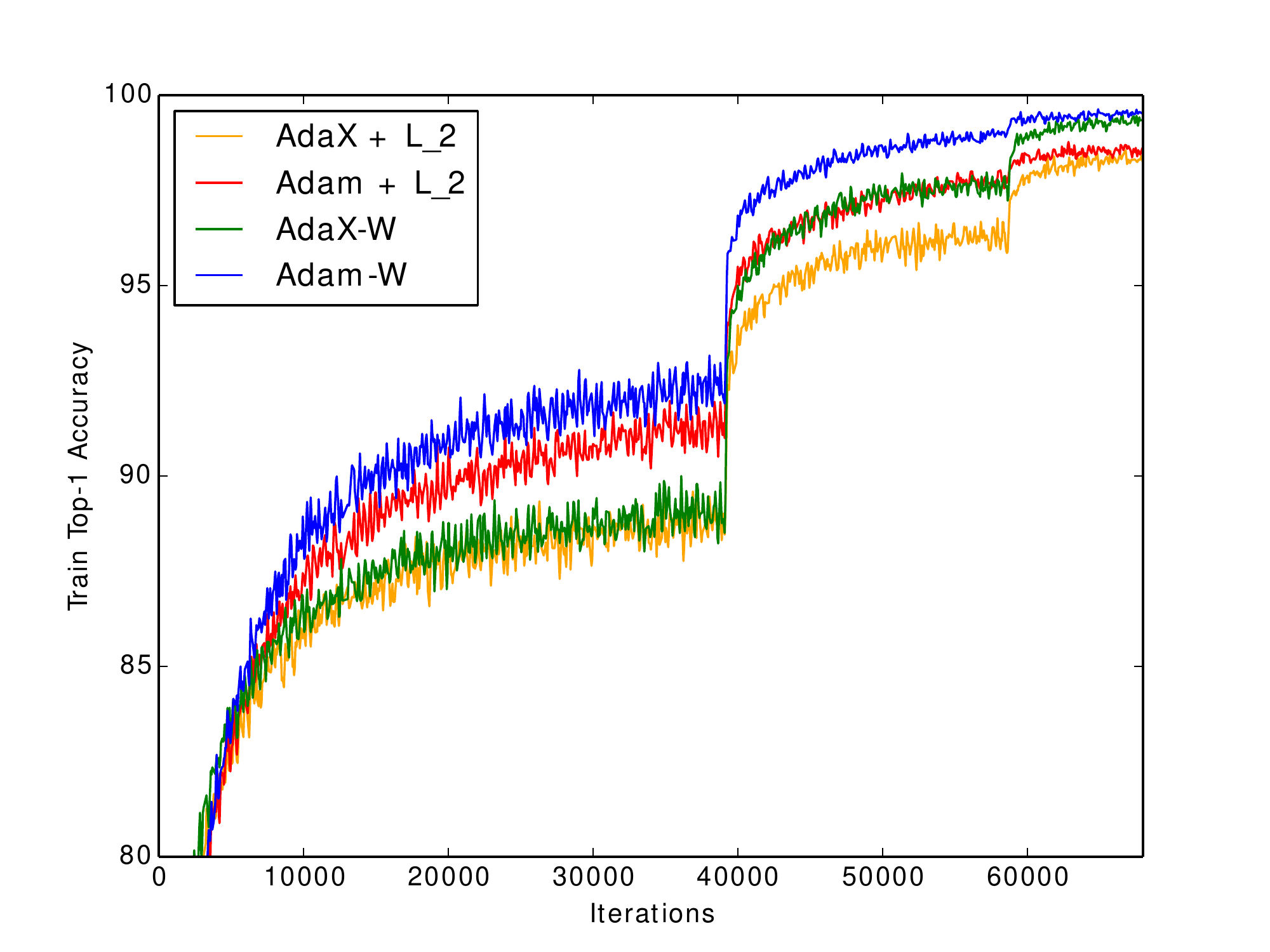}
}
\subfigure[Testing Top-1 Accuracy]{
  \centering
  \includegraphics[width=0.4\linewidth]{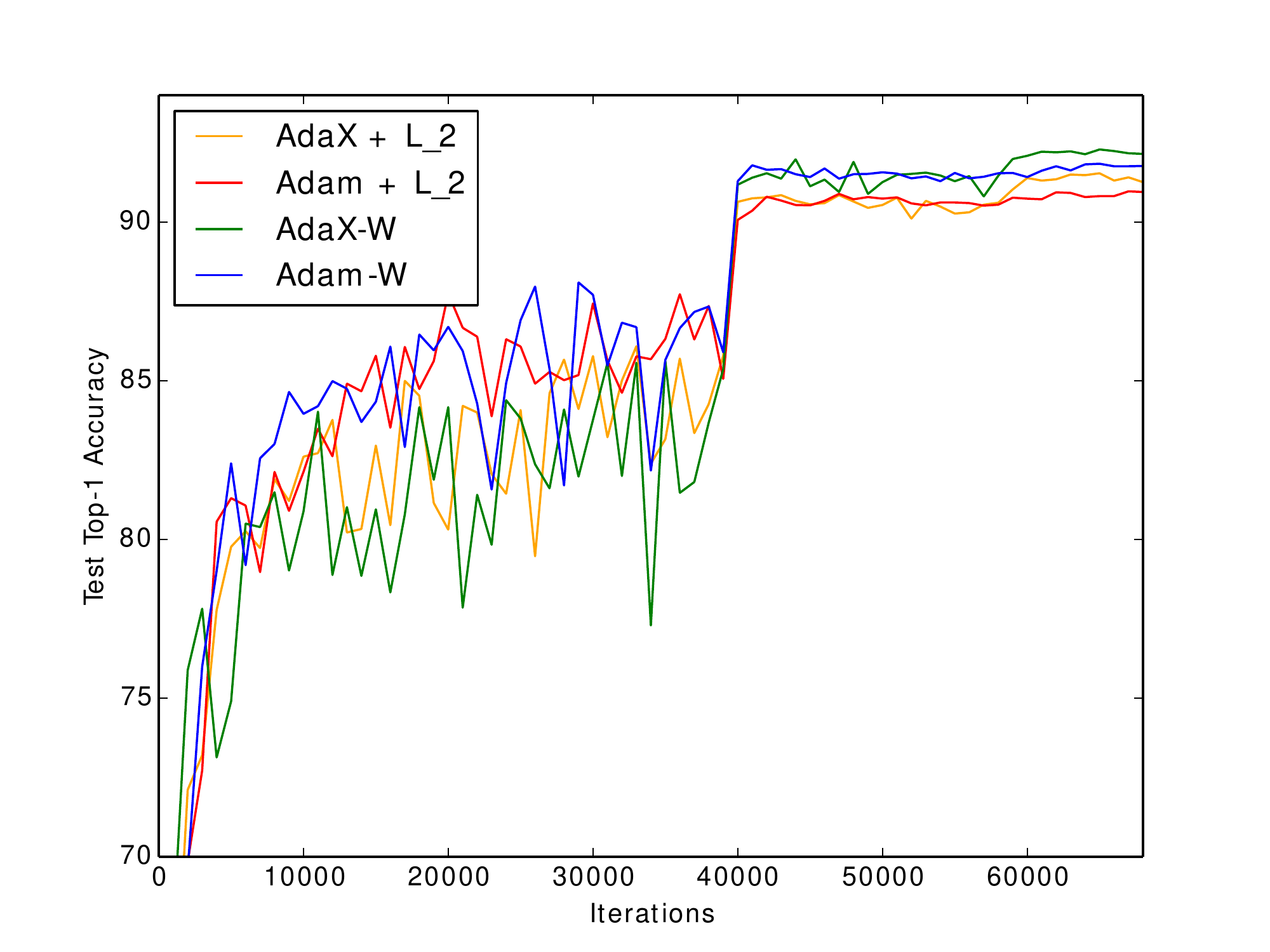}
}
\caption{Training and Testing Results on CIFAR-10 with $L_2$ Regularization and Weight Decay. AdaX performed better than Adam both when using $L_2$ regularization and decoupled weight decay}
\label{fig: L2-WD}
\end{figure}

\subsection{Hyper-parameter Tuning and Experimental details}
\label{App: Parameter_Tuning}
The hyperparameters in different algorithms have a huge impact on their performances in the experiments. To find the optimal hyperparameters that could generate the best results, we thoroughly tuned the hyperparameters in the algorithms.

\textbf{Step size}. We followed \citet{Wilson2017The} to perform a logarithmically-spaced grid search of the optimal step sizes and the step sizes we tried were listed in the following tables, where the step sizes in bold were the ones with best performances and used in the experiments section.

\textbf{Step size: Image Classification (CIFAR, ImageNet)}
\begin{itemize}
  \setlength\itemsep{-0.5em}
  \item \textbf{SGD(M)}  \{10, 1, \textbf{1e-1}, 1e-2, 1e-3\} \\
  \item \textbf{AdamW}  \{1e-2, 3e-3, \textbf{1e-3}, 3e-4, 1e-4\}\\
  \item \textbf{AMSGrad}  \{1e-2, 3e-3, \textbf{1e-3}, 3e-4, 1e-4\}\\
  \item \textbf{AdaX-W(ours)}  \{1e-2,\textbf{5e-3}, 4e-3, 3e-3, {2.5e-3}, 1e-3, 1e-4, 5e-5, 1e-5\}\\
\end{itemize}
\textbf{Step size: VOC2012 Segmentation}
\begin{itemize}
  \setlength\itemsep{-0.5em}
  \item \textbf{SGD(M)}  \{1e-3, 5e-4, \textbf{2.5e-4}, 1e-4, 5e-5\} \\
  \item \textbf{AdamW}  \{5e-4, 1e-5, 5e-5, \textbf{1e-6}, 5e-7\}\\
  \item \textbf{AMSGrad}  \{5e-4, 1e-5, 5e-5, \textbf{1e-6}, 5e-7\}\\
  \item \textbf{AdaX-W(ours)}  \{1e-4, {5e-5}, \textbf{1e-5}, 5e-6, 1e-6\}\\
\end{itemize}

\textbf{Step size: Billionwords}
\begin{itemize}
  \setlength\itemsep{-0.5em}
  \item \textbf{Adam, AMSGrad}  \{5e-3, 2e-3, 1e-3, \textbf{5e-4}, 1e-4 \}\\

  \item \textbf{AdaX(ours)}  \{5e-3, 2e-3, 1e-3, \textbf{5e-4}, 1e-4\} $\times$ \{0.5, 1, 5, \textbf{15}, 25, 50, 100\}  (best 7.5e-3) \\
\end{itemize}

\begin{figure}[H]
\centering
\subfigure[Training Top-1 Accuracy]{
  \centering
  \includegraphics[width=0.4\linewidth]{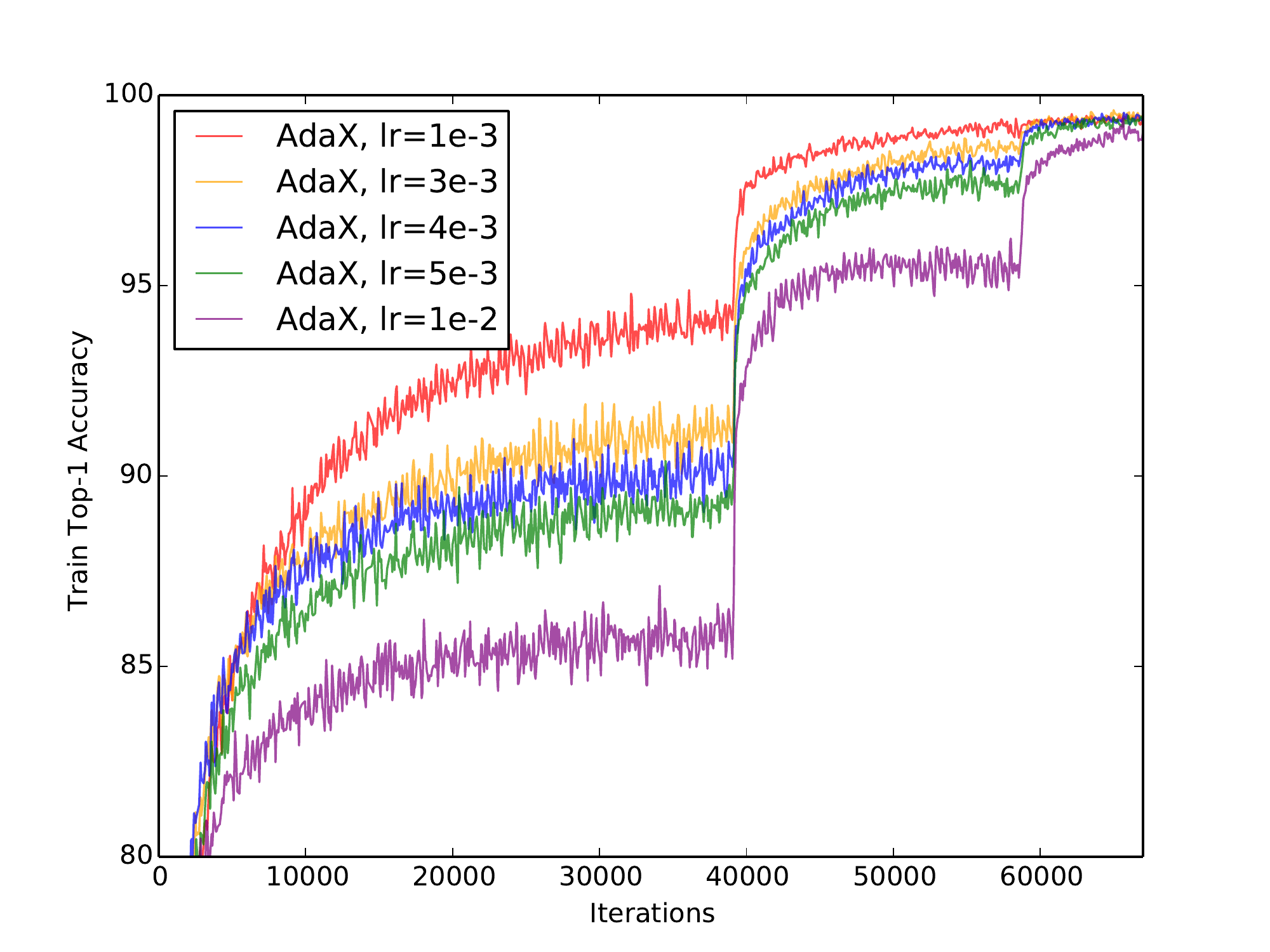}
}
\subfigure[Testing Top-1 Accuracy]{
  \centering
  \includegraphics[width=0.4\linewidth]{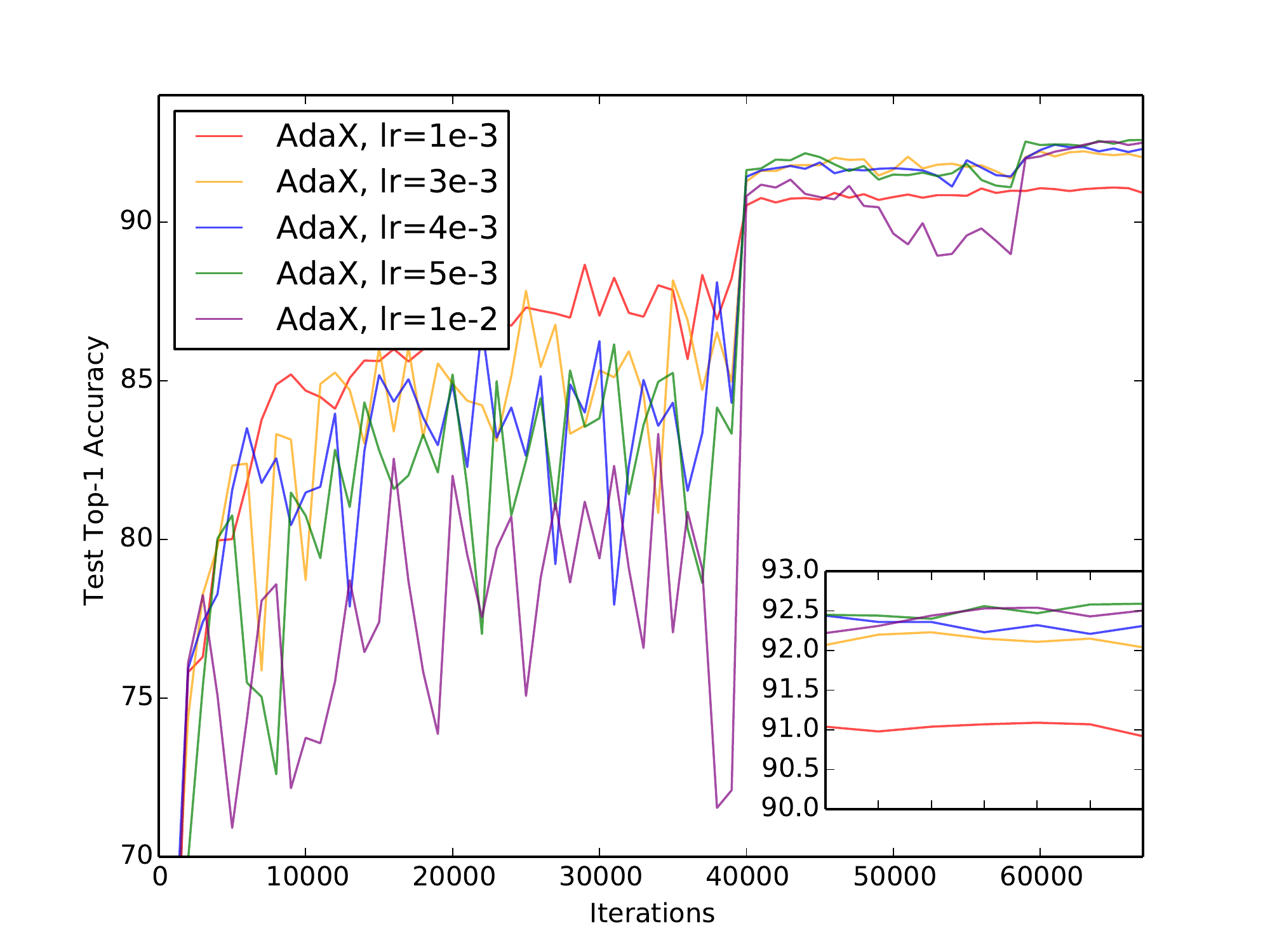}
}
\caption{Training and Testing Results on CIFAR-10 using AdaX-W with different initial step sizes. It can be observed from the figure that AdaX is robust to small step size changes. However, very small step sizes ($lr=1e-3$) can lead to over-fitting and bad testing results.}
\label{fig: AdaX-lr}
\end{figure}

\textbf{Momentum parameters.} For the momentum parameters of AdamW and AMSGrad, we tuned over $(\beta_1,\beta_2) = \{(0.9, 0.999), (0.99, 0.999),(0.99, 0.9999)\}$ and found that the default values $(0.9, 0.999)$ as in \citet{Kingma2015Adam} yielded the best result. For the momentum parameters $(\beta_1, \beta_2)$ in AdaX, we directly applied $\beta_1 = 0.9$ as in Adam and we tuned $\beta_2$ over $\{1e-3, 1e-4, 1e-5\}$. As shown in Figure \ref{fig: AdaX-beta2}, we found that the value of $\beta_2$ didn't affect the general performance of AdaX, which again proves our claim that AdaX's second moment is more stable than Adam's. A default value of $1e-4$ was applied in all of our experiments.

\begin{figure}
\centering
\subfigure[Training Top-1 Accuracy]{
  \centering
  \includegraphics[width=0.4\linewidth]{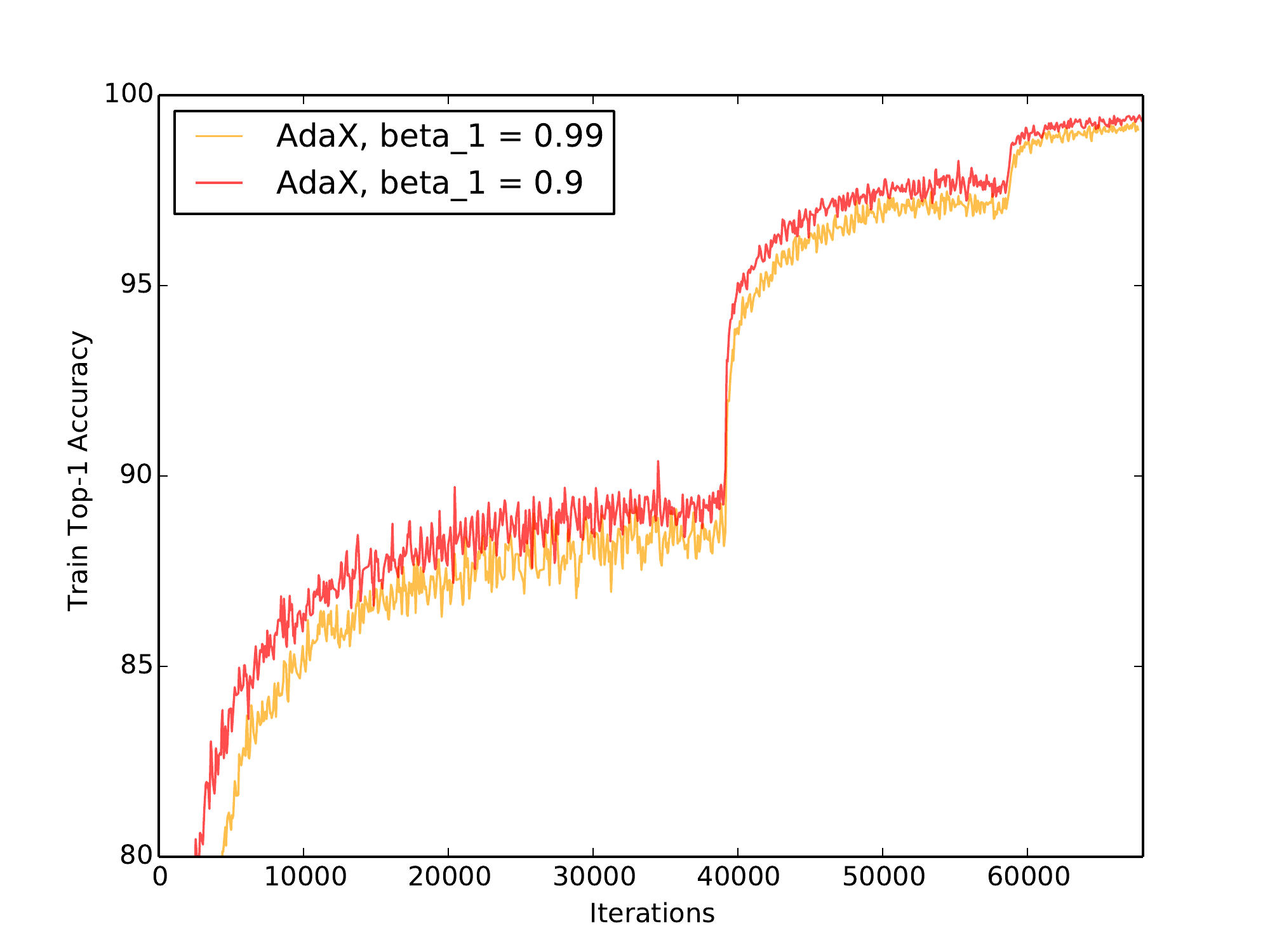}
}
\subfigure[Testing Top-1 Accuracy]{
  \centering
  \includegraphics[width=0.4\linewidth]{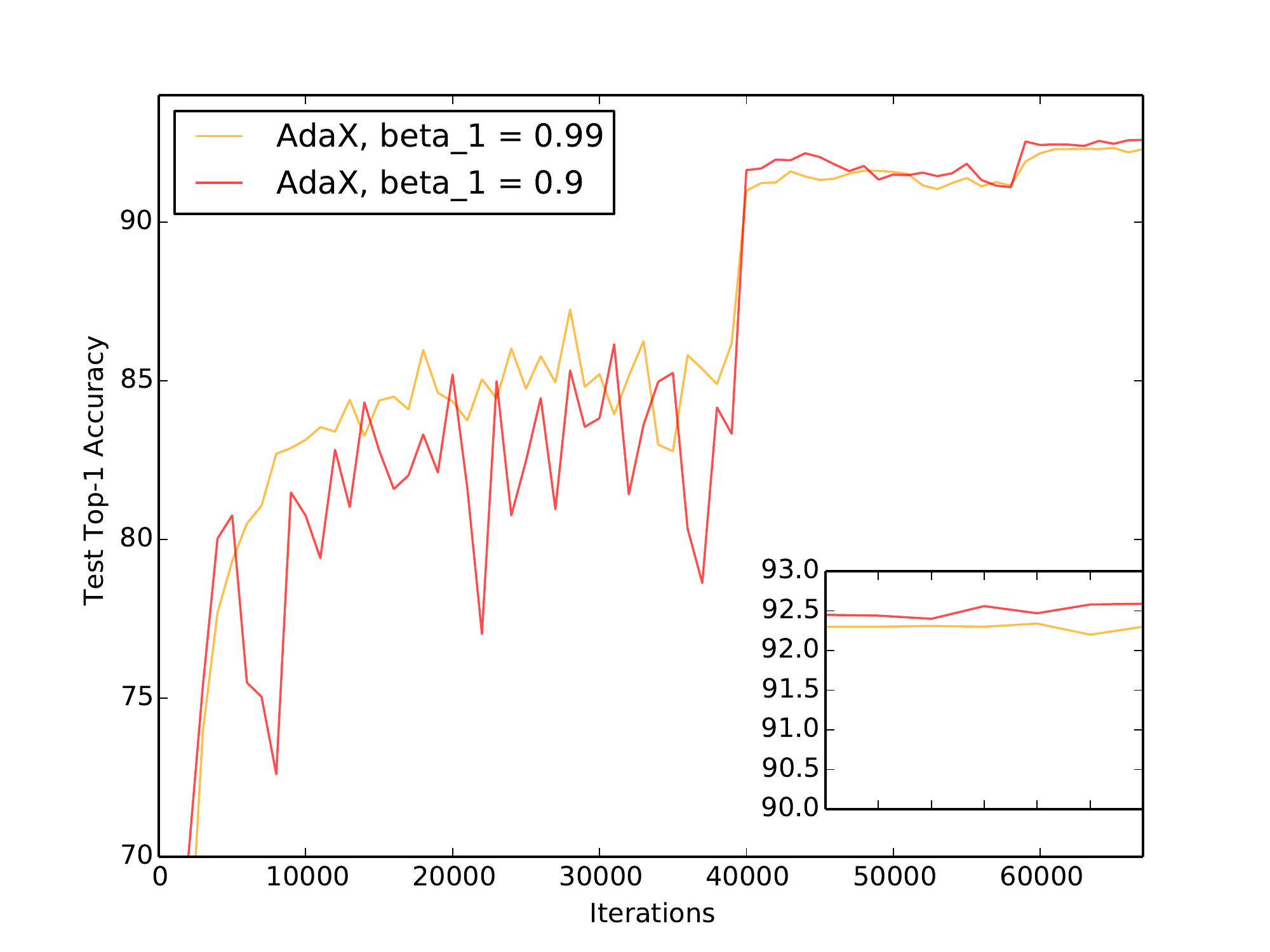}
}
\caption{Training and Testing Results on CIFAR-10 using AdaX-W with different $\beta_1$'s. Setting $\beta_1 = 0.9$ as in Adam yields the best results in our experiments}
\label{fig: AdaX-beta1}
\end{figure}

\begin{figure}
\centering
\subfigure[Training Top-1 Accuracy]{
  \centering
  \includegraphics[width=0.4\linewidth]{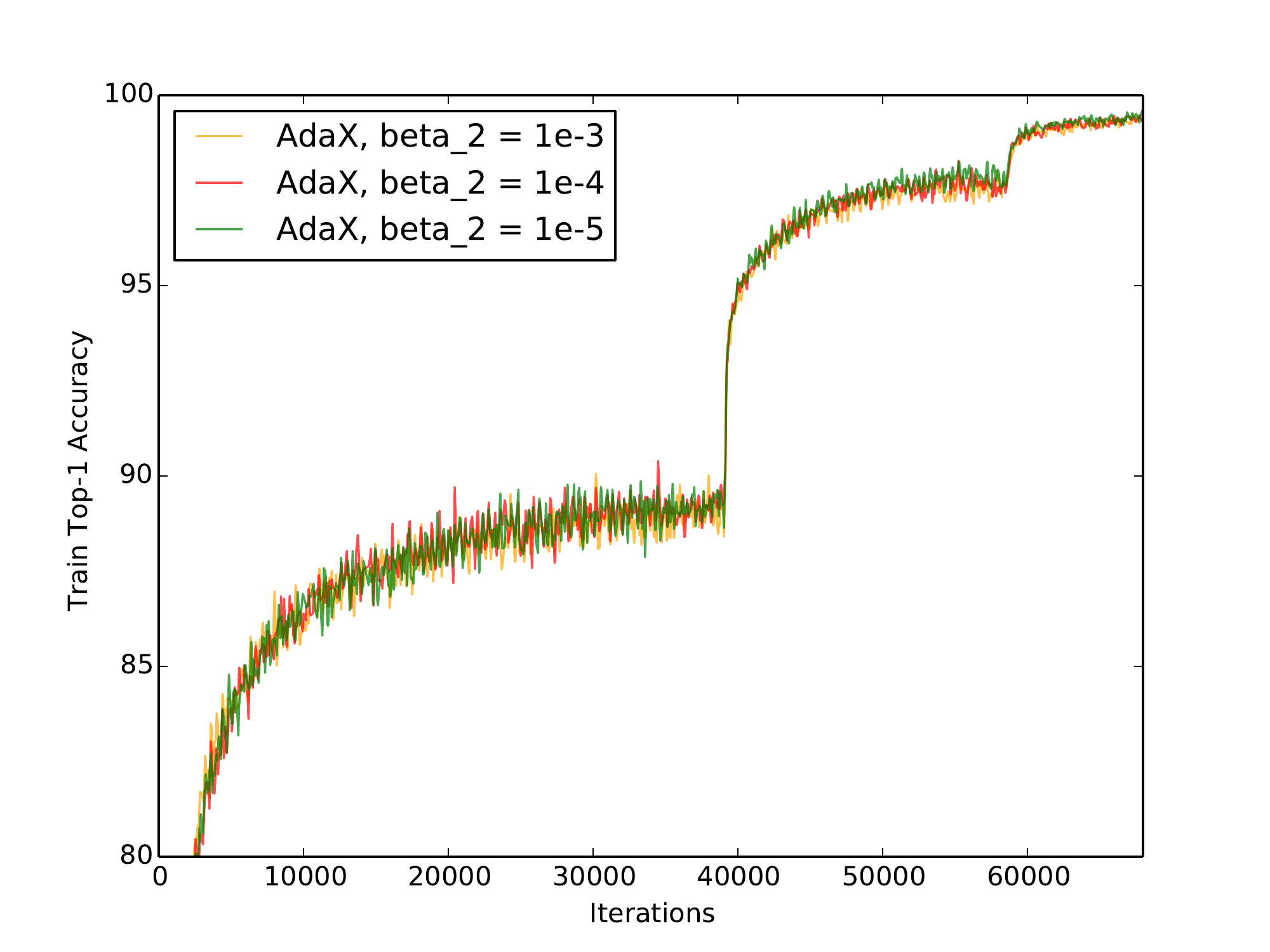}
}
\subfigure[Testing Top-1 Accuracy]{
  \centering
  \includegraphics[width=0.4\linewidth]{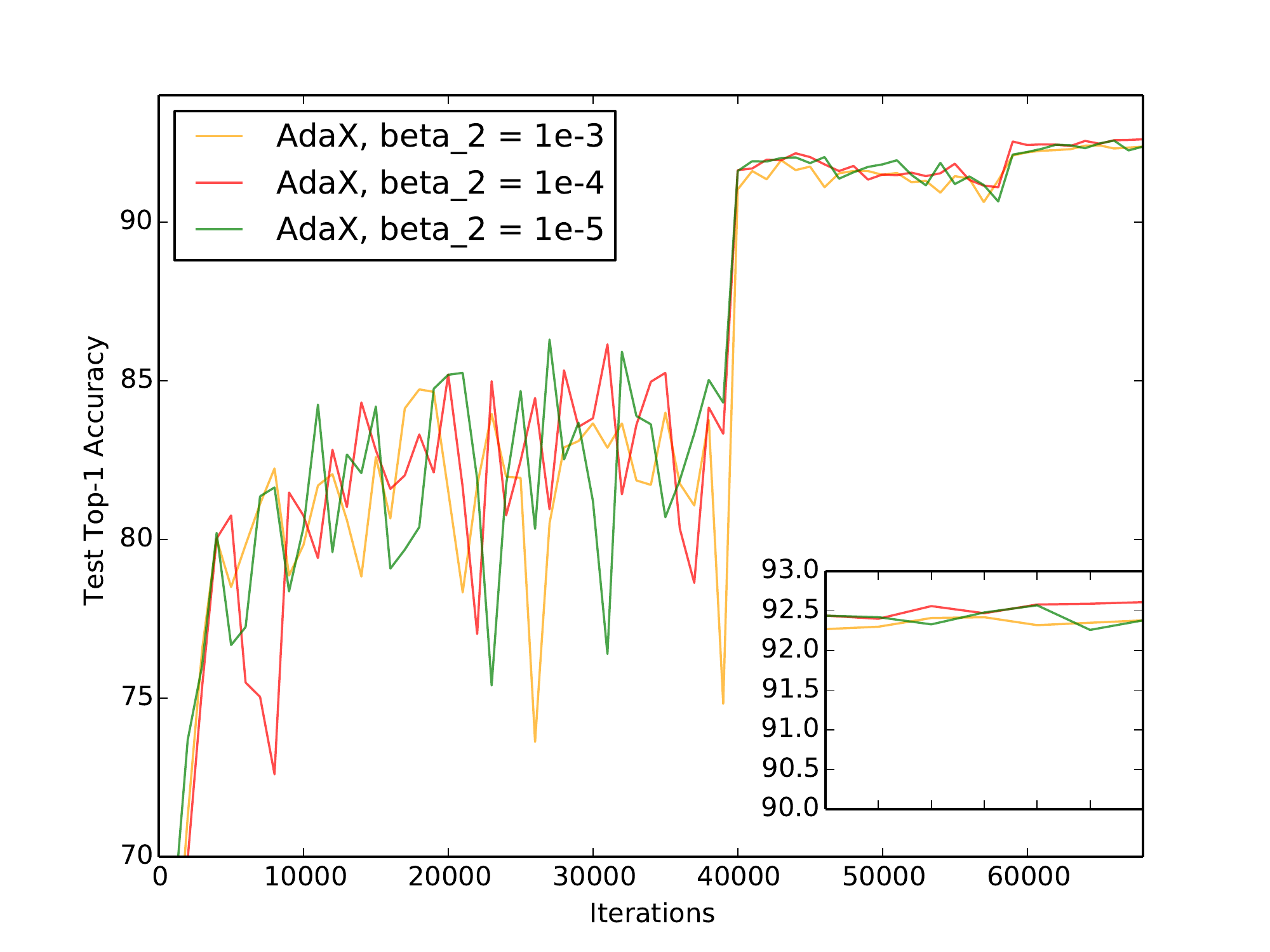}
}
\caption{Training and Testing Results on CIFAR-10 using AdaX-W with different $\beta_2$. The results show that AdaX is not sensitive to the value of $\beta_2$. However, we recommend using smaller ones since larger $\beta_2$ induces higher computational costs. The default $\beta_2$ is 1e-4 in all our experiments.}
\label{fig: AdaX-beta2}
\end{figure}

\textbf{Weight decay}. For SGDM, the same weight decays as in the original papers \cite{He2016Deep}\cite{Chen2016Deeplab}\cite{rdspring1} \cite{Loshchilov2019Decoupled} were used in all the experiments. For AdamW and AMSGrad, we tuned over \{5e-1, 1e-1, 5e-2, 1e-2\} and found that 1e-1 generated fast convergence as well as good performances. For AdaX-W, we directly applied 5e-2 weight decay for all our experiments on CIFAR-10, ImageNet and VOC2012 Segmentation. For the One-Billion Word dataset, 0 weight decay was applied for all the algorithms.

\textbf{Batch size.} The mini-batch sizes used in our experiments were the same as those in the original papers, i.e. 128 for CIFAR-10 and 256 (8 GPUs) for ImageNet as in \citet{He2016Deep}, 10 for VOC2012 as in \citet{Chen2016Deeplab}, and 128 for One-Billion word as in \citet{rdspring1}. 

For the other hyper-parameters such as dropout probability, we directly applied the same settings as in the baselines \cite{He2016Deep}\cite{Chen2016Deeplab}\cite{rdspring1} \cite{Loshchilov2019Decoupled}.

\begin{table}
\caption{\footnotesize Comparisons of Runtime. All our experiments were conducted using Nvidia-Tesla V100 GPUs. We reported the average runtime over 5 independent runs. (h: hours)}
\centering
\small
\begin{tabular}[t]{ c|c| c| c |c}
\hline
 Method & CIFAR& ImageNet (ResNet-18)&VOC 2012 & One Billion Word\\
 \hline
AdamW & 0.36h & 50.10h & 9.10h &95.58h \\
\textbf{AdaX-W(ours)}& 0.35h & 50.74h  & 8.89h & 96.82h\\
\hline
\end{tabular}
\label{tab: time}
\end{table}

\subsection{More Experiments on ImageNet}

We also conducted more experiments on ImageNet with a much larger neural network, ResNet-50 \citep{He2016Deep} and reported the training curve and the final accuracy in Figure \ref{fig: IN_res50} and Table \ref{tab: IN_res50}. We used a cosine learning rate scheduler \citep{loshchilov2016sgdr} which decreased the step sizes with a cosine curve to 1e-6 at the end. Similarly, a warm up scheme was also applied in the initial 25k iterations \cite{Goyal2017Accurate}. The initial step sizes and the hyper-parameters were exactly the same as in section \ref{sec: Experiments}. As can be observed in the figures and the table, AdamW performed poorly on this task and its final accuracy was much lower than SGDM's. AMSGrad did not improve AdamW by a significant margin. Although our method did not catch up with SGDM at the end, its performance was much better than AdamW and it had fast convergence as well as much higher training accuracy. 

\begin{figure}[H]
\centering
\subfigure[Training Top-1 Accuracy]{
  \centering
  \includegraphics[width=0.4\linewidth]{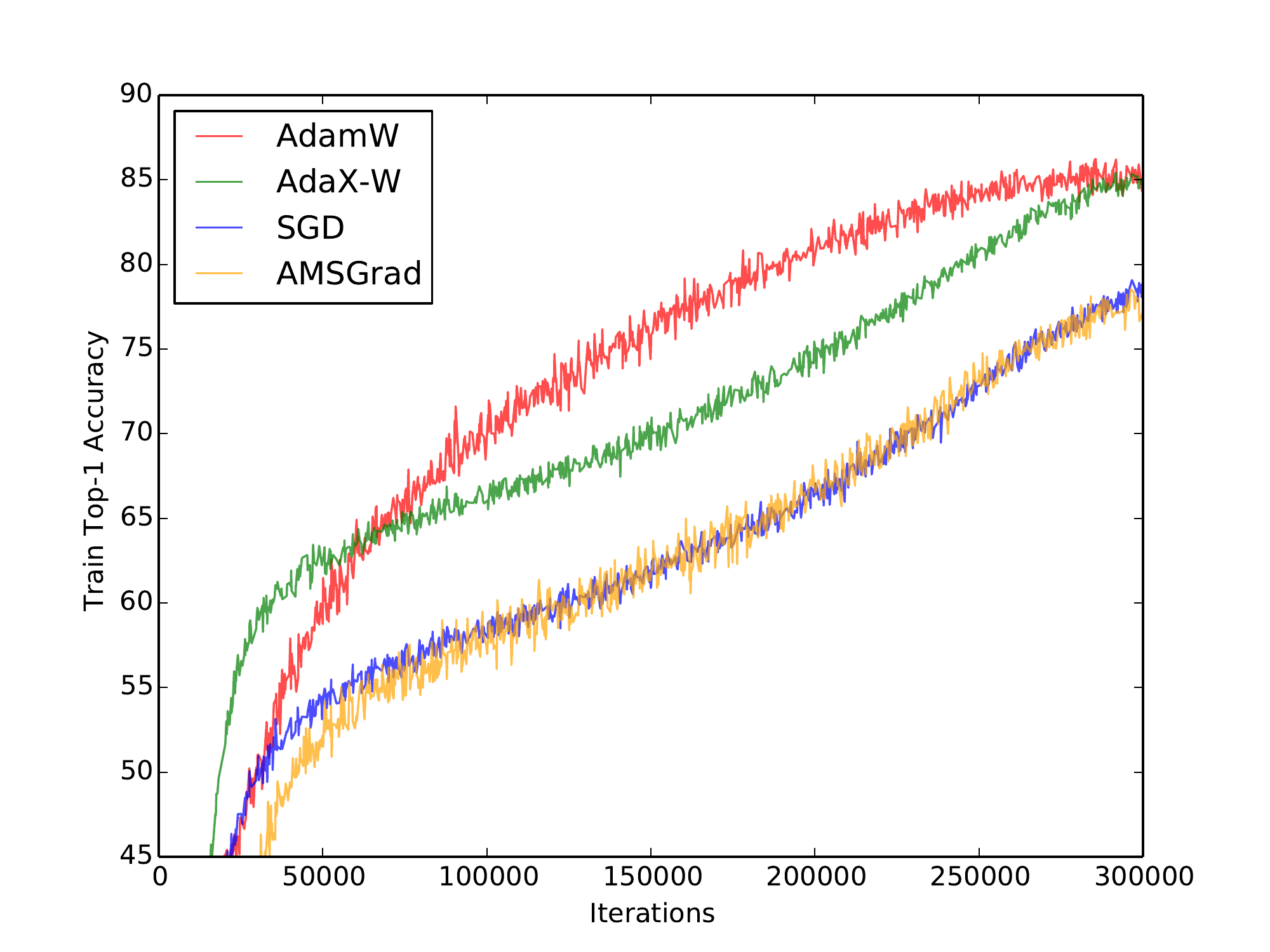}
}
\subfigure[Testing Top-1 Accuracy]{
  \centering
  \includegraphics[width=0.4\linewidth]{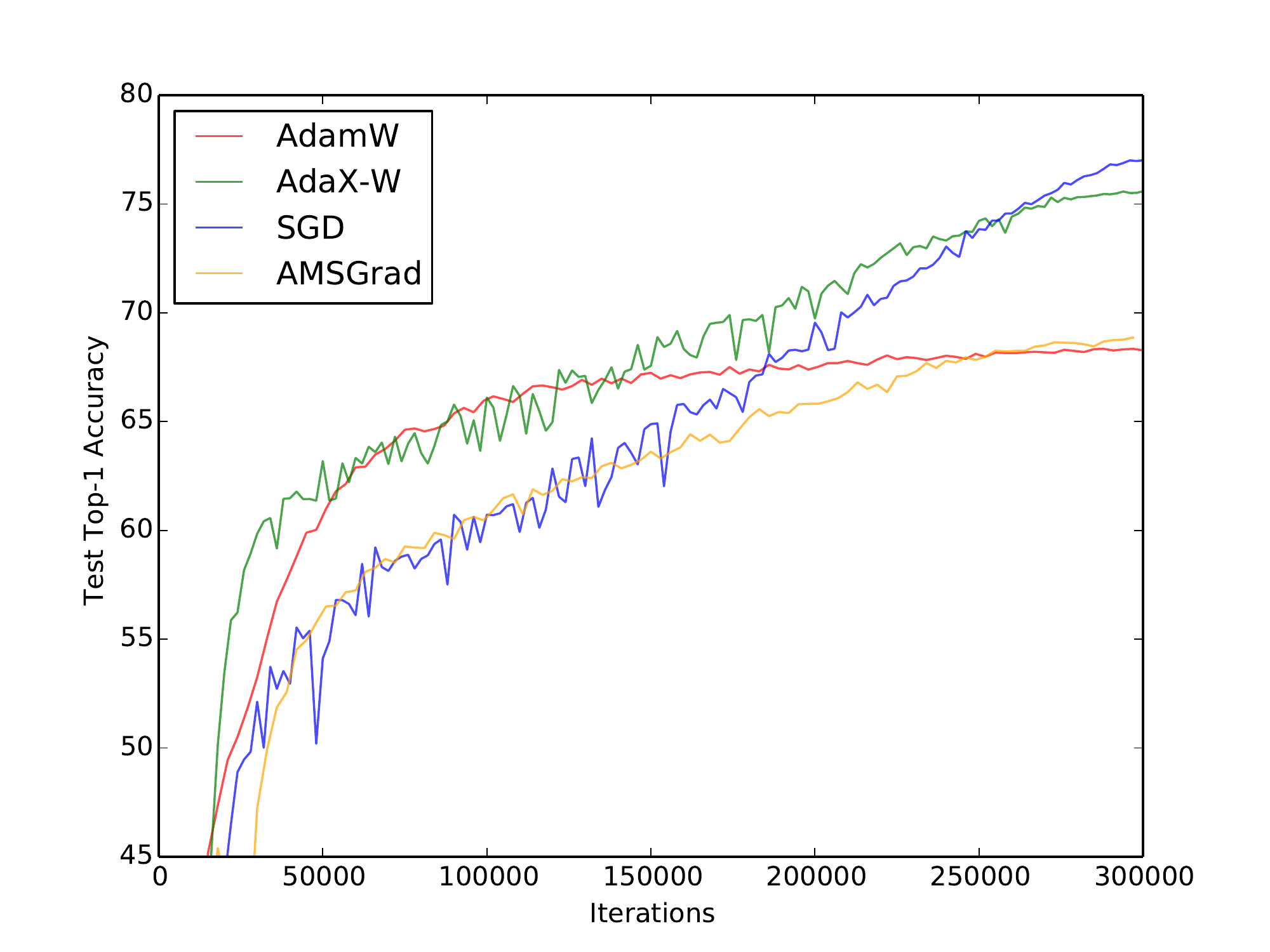}
}
\caption{Training and Testing Results on ImageNet using ResNet-50}
\label{fig: IN_res50}
\end{figure}

\begin{table}[H]
\caption{\footnotesize Validation Top-1 accuracy on ImageNet with ResNet-50. We reported the results average over 5 independent runs and the margin of error.}
\centering
\small
\begin{tabular}[t]{ c|c}
\hline
 Method & Top-1 Accuracy\\
 \hline
\textbf{SGDM }& \textbf{77.12 $\pm$ 0.07}\\
AdamW & 68.27 $\pm$ 0.08  \\
AMSGrad(W) & 68.76 $\pm$ 0.12\\
\textbf{AdaX-W(ours)}&\textbf{ 75.58 $\pm$ 0.08} \\

\hline
\end{tabular}
\label{tab: IN_res50}
\end{table}

\end{section}

\end{document}